\documentclass[journal]{IEEEtran}

\newif\ifdraft

\drafttrue

\usepackage{silence}

\usepackage[T1]{fontenc}

\usepackage[english]{babel}

\usepackage{csquotes}

\usepackage{xparse}

\usepackage{etoolbox}

\usepackage[dvipsnames,table]{xcolor}

\usepackage{amsmath}

\usepackage{amssymb}

\usepackage{amsfonts}

\usepackage{mathrsfs}

\usepackage{tabularray}

\UseTblrLibrary{booktabs}

\usepackage{graphicx}

\usepackage[export]{adjustbox}

\usepackage{subcaption}

\usepackage{tikz}

\usepackage{pgfplots}

\pgfplotsset{compat=1.18}

\usetikzlibrary{
  arrows.meta,
  backgrounds,
  positioning,
  shapes
}
\usepackage{epstopdf}
\newif\ifusealgorithmtwoe

\usealgorithmtwoefalse

\ifusealgorithmtwoe

  \usepackage[
    ruled,
    vlined,
    linesnumbered
  ]{algorithm2e}

  \SetAlFnt{\small}
  \SetAlCapFnt{\small}
  \SetAlCapNameFnt{\small}
  \SetAlgoNlRelativeSize{-1}

  \SetKwInput{Input}{Input}
  \SetKwInput{Output}{Output}
  \SetKwInput{Precondition}{Precondition}
  \SetKwInput{Postcondition}{Postcondition}

\else

  \usepackage{algorithm}

  \usepackage[noend]{algpseudocode}

  \algrenewcommand\algorithmicrequire{\textbf{Precondition:}}
  \algrenewcommand\algorithmicensure{\textbf{Postcondition:}}

  \algnewcommand\algorithmicinput{\textbf{Input:}}
  \algnewcommand\Input{\item[\algorithmicinput]}

  \algnewcommand\algorithmicoutput{\textbf{Output:}}
  \algnewcommand\Output{\item[\algorithmicoutput]}

  \algnewcommand\algorithmicforeach{\textbf{for each}}
  \algdef{S}[FOR]{ForEach}[1]{\algorithmicforeach\ #1\ \algorithmicdo}

  \algdef{SE}[IF]{IfNot}{EndIf}[1]
    {\algorithmicif\ \algorithmicnot\ #1\ \algorithmicthen}
    {\algorithmicend\ \algorithmicif}

\fi

\usepackage[
  style=ieee,
  backend=biber,
  dashed=false,
  doi=false,
  isbn=false,
  url=false,
  eprint=false,
  sorting=none
]{biblatex}

\makeatletter
\def\abx@missing@entry#1{\abx@missing{??}}
\makeatother

\usepackage{newtxtext,newtxmath}

\usepackage{siunitx}

\usepackage[hidelinks]{hyperref}

\usepackage{orcidlink}

\usepackage{stfloats}

\usepackage{pifont}

\ifdraft

  \usepackage[
    colorinlistoftodos,
    prependcaption,
    textsize=footnotesize
  ]{todonotes}

\else

  \usepackage[disable]{todonotes}

\fi

\newcounter{todocounter}

\NewDocumentCommand{\unsure}{O{inline} m}{%
  \stepcounter{todocounter}%
  \todo[
    linecolor=red,
    backgroundcolor=red!25,
    bordercolor=red,
    #1
  ]{\thetodocounter: Unsure: #2}%
}

\NewDocumentCommand{\change}{O{inline} m}{%
  \stepcounter{todocounter}%
  \todo[
    linecolor=blue,
    backgroundcolor=blue!25,
    bordercolor=blue,
    #1
  ]{\thetodocounter: Change: #2}%
}

\NewDocumentCommand{\info}{O{inline} m}{%
  \stepcounter{todocounter}%
  \todo[
    linecolor=OliveGreen,
    backgroundcolor=OliveGreen!25,
    bordercolor=OliveGreen,
    #1
  ]{\thetodocounter: Info: #2}%
}

\NewDocumentCommand{\improvement}{O{inline} m}{%
  \stepcounter{todocounter}%
  \todo[
    linecolor=Plum,
    backgroundcolor=Plum!25,
    bordercolor=Plum,
    #1
  ]{\thetodocounter: Improvement: #2}%
}

\NewDocumentCommand{\roletodo}{O{inline} m m m}{%
  \stepcounter{todocounter}%
  \todo[
    linecolor=#2,
    backgroundcolor=#2!20,
    bordercolor=#2,
    #1
  ]{\thetodocounter: #3: #4}%
}

\NewDocumentCommand{\nicktodo}{O{inline} m}{%
  \roletodo[#1]{RoyalBlue}{Nick}{#2}%
}

\NewDocumentCommand{\krishnatodo}{O{inline} m}{%
  \roletodo[#1]{Violet}{Krishna}{#2}%
}

\NewDocumentCommand{\studenttodo}{O{inline} m}{%
  \roletodo[#1]{Orange}{Student}{#2}%
}

\NewDocumentCommand{\thiswillnotshow}{O{} m}{%
  \stepcounter{todocounter}%
  \todo[disable,#1]{\thetodocounter: #2}%
}

\usepackage{enumitem}

\usepackage{adjustbox}
\usepackage{amsmath,amssymb}
\usepackage{newtxtext,newtxmath}
\usepackage{xcolor}
\usepackage{tikz}
\usetikzlibrary{arrows.meta,positioning,calc,fit,backgrounds,shapes.geometric}
\usepackage{fontawesome5}
\usepackage{graphicx}
\usepackage{twemojis}
\definecolor{memblue}{RGB}{0,45,255}
\definecolor{physorange}{RGB}{255,80,0}
\definecolor{respurple}{RGB}{90,20,180}
\definecolor{softblue}{RGB}{245,250,255}
\definecolor{softorange}{RGB}{255,250,242}
\definecolor{softpurple}{RGB}{250,247,255}

\robustify
\robustify\addtocounter
\robustify
\robustify

\NewDocumentCommand{\method}{}{\textsc{DynSSM}}

\NewDocumentCommand{\ddm}{}{\textsc{DDM}}
\NewDocumentCommand{\fthd}{}{\textsc{FTHD}}

\NewDocumentCommand{\sota}{}{\textsc{SoTA}}

\NewDocumentCommand{\figref}{m}{Fig.~\ref{#1}}

\NewDocumentCommand{\vx}{}{\ensuremath{v_x}}

\NewDocumentCommand{\vy}{}{\ensuremath{v_y}}

\NewDocumentCommand{\yawrate}{}{\ensuremath{\omega}}
\NewDocumentCommand{\w}{}{\yawrate}

\NewDocumentCommand{\af}{}{\ensuremath{\alpha_f}}
\NewDocumentCommand{\ar}{}{\ensuremath{\alpha_r}}

\NewDocumentCommand{\Ffz}{}{\ensuremath{F_{fz}}}
\NewDocumentCommand{\Frz}{}{\ensuremath{F_{rz}}}

\NewDocumentCommand{\Frx}{}{\ensuremath{F_{rx}}}

\NewDocumentCommand{\Ffy}{}{\ensuremath{F_{fy}}}
\NewDocumentCommand{\Fry}{}{\ensuremath{F_{ry}}}

\NewDocumentCommand{\SD}{}{\ensuremath{\mathcal{S}_{\mathcal{D}}}}
\NewDocumentCommand{\SDtr}{}{\ensuremath{\mathcal{S}_{\mathcal{D}}^{\mathrm{train}}}}
\NewDocumentCommand{\SDte}{}{\ensuremath{\mathcal{S}_{\mathcal{D}}^{\mathrm{test}}}}

\NewDocumentCommand{\RD}{}{\ensuremath{\mathcal{R}_{\mathcal{D}}}}
\NewDocumentCommand{\RDtr}{}{\ensuremath{\mathcal{R}_{\mathcal{D}}^{\mathrm{train}}}}
\NewDocumentCommand{\RDte}{}{\ensuremath{\mathcal{R}_{\mathcal{D}}^{\mathrm{test}}}}

\NewDocumentCommand{\cmark}{}{\ding{51}}
\NewDocumentCommand{\xmark}{}{\ding{55}}

\title{DynSSM: A Physics-Aware State-Space Memory Framework for Learning Vehicle Dynamics}

\author{Musabbir Ahmed Arrafi \IEEEmembership{Student Member, IEEE}, Malik Ali \IEEEmembership{Student Member, IEEE},\\%
Nicholas M. Stiffler \IEEEmembership{Member, IEEE},
and Krishna Bhavithavya Kidambi \IEEEmembership{Senior Member, IEEE}%
\thanks{The authors are with the University of Dayton, Dayton, OH 45469 USA.
M. A. Arrafi and K. B. Kidambi are with Mechanical and Aerospace Engineering; M. Ali is with Electrical and Computer Engineering;
and N. M. Stiffler is with Computer Science.
E-mail: arrafim1@udayton.edu, alim37@udayton.edu, nstiffler1@udayton.edu, kkidambi1@udayton.edu.}}

\begin{document}

\maketitle


\begin{abstract}
Accurate modeling of nonlinear vehicle dynamics is essential for high-speed autonomous racing, where controllers
operate at the handling limits. 
Model-based methods are interpretable but rely on simplifying assumptions, while purely learned models capture nonlinearities yet often lack physical consistency, generalization, and adaptability to changing operating conditions.
This paper presents DynSSM, a physics-aware state-space memory framework that combines learned temporal representations with a structured vehicle dynamics model.
The proposed approach integrates state-space sequence modeling and recurrent encoders to capture long- and short-term dynamic behavior. It simultaneously adapts tire and vehicle-dynamics parameters within bounded ranges to preserve physical plausibility.
A residual correction mechanism compensates for remaining unmodeled dynamics while preserving the underlying physics-based structure.
DynSSM is evaluated on both simulated small-scale racing data and real-world full-scale autonomous Indy racecar data. 
When evaluated on an unseen real-world track, DynSSM reduces one-step prediction RMSE by up to $27.2\%$ in longitudinal velocity, $73.9\%$ in lateral velocity, and $88.8\%$ in yaw rate compared with the state-of-the-art (\sota{}) baselines.
Component-wise ablation studies demonstrate the importance of temporal memory, parameter adaptation, and residual correction for predictive performance and robustness. 
Further, closed-loop simulations using nonlinear model predictive control demonstrate that DynSSM remains feasible across the evaluated tracks and achieves up to a 13.2\% reduction in one-lap completion time compared with \sota{} baselines.
These results indicate that combining temporal memory, bounded physics-guided parameter adaptation, and residual correction provides an accurate, interpretable, and control-ready dynamics model for autonomous racing.
\end{abstract}




\section{Introduction} \label{sec:intro}
Autonomous racing at high speed has emerged as a demanding testbed for developing advances in control, planning, and vehicle  dynamics ~\cite{moon2024autonomous,thakkar2024hierarchical}.
More recently, the racing domain has also been used to develop and evaluate learning-based methods capable of operating under aggressive vehicle maneuvers and rapidly changing driving conditions~\cite{zhou2025adaptive,dikici2025learning}.
Unlike conventional autonomous driving, racing vehicles operate close to the limits of tire adhesion, actuator authority, and vehicle stability, where sharp cornering, lateral-load variation, rapid high-G braking and acceleration produce strongly nonlinear and history-dependent behavior~\cite{musiu2026comprehensive, singh2019literature}. 
Accurate, robust, and efficient dynamics modeling is essential for predictive control, trajectory optimization, and real-time decision-making \cite{zhang2024survey}. 
At the same time, such models must remain lightweight enough for repeated online evaluation within closed-loop controllers operating under strict real-time constraints \cite{li2025autonomous, baumann2025forzaeth, xu2026small}.
Consequently, autonomous racing exposes a fundamental modeling challenge: developing dynamics representations that are accurate, temporally aware, physically interpretable, and computationally efficient.
Beyond racing itself, insights gained here directly transfer to broader autonomous driving applications where reliability and safety margins are critical.

Estimating vehicle dynamics for autonomous racing has been studied across scaled platforms, full-scale racecars, and simulation-based control benchmarks \cite{cheng2025deep, leanza2024vehicle}. 
Existing approaches can be broadly grouped into four categories: 
(i) physics-based vehicle dynamics models, 
(ii) data-driven neural dynamics models, 
(iii) hybrid physics-aware learning architectures, and 
(iv) state-space and recurrent sequence models for dynamical systems \cite{guiggiani2018science}.

Classical physics-based vehicle dynamics models like kinematic, single-track and double-track describe vehicle motion using analytical relationships derived from Newtonian mechanics. 
Kinematic models are widely used in planning and control because of their simplicity and low computational cost, but they neglect inertial coupling, tire slip, and friction-limited behavior~\cite{wang2021path,bellegarda2021dynamic}. 
Dynamic single-track and double-track models improve fidelity by incorporating lateral and longitudinal tire forces, yaw dynamics, and slip-angle effects~\cite{wischnewski2022tube}. 
Higher-order multi-degree-of-freedom models can further represent suspension motion, load transfer, and coupled tire-road interaction, but they require careful parameter identification and incur greater computational cost~\cite{zhang2024survey}. 
Pacejka-style tire models provide a powerful representation of nonlinear tire forces, yet their accuracy depends on calibrated coefficients that may vary with load, tire condition, surface, and operating regime~\cite{pacejka2012tire}. 
These limitations become more pronounced in autonomous racing, where aggressive acceleration, braking, and cornering repeatedly push the vehicle into high-slip and transient regimes~\cite{rucco2015minimum}. 
Thus, physics-based models offer interpretability and control compatibility, but 
often assume fixed tire parameters and quasi-static load behavior, which can break down during high-slip maneuvers, and changing tire-road interaction encountered during racing.

Data-driven models seek to reduce reliance on manual parameter identification
offer an alternative by learning nonlinear dynamics directly from telemetry~\cite{spielberg2019neural,becker2023model}.
Neural-network models have been used to approximate nonlinear vehicle dynamics and can outperform simplified analytical models when evaluated on data drawn from similar operating conditions~\cite{badgujar2023deep}. 
Learning-based approaches have also been explored for aggressive driving and racing, including differentiable filtering with learned dynamics and online friction estimation~\cite{wkegrzynowski2024learning}. 
These methods are attractive because they can capture complex nonlinear behavior that is difficult to model analytically. 
However, purely black-box predictors provide limited physical interpretability, may extrapolate poorly outside the training distribution and provide infeasible predictions that destabilize closed-loop control \cite{kim2022physics}.
This is especially problematic in model-based control, where the learned model is recursively queried over a prediction horizon and small errors can accumulate into infeasible trajectories or unstable control actions. 
End-to-end racing policies, such as DeepRacing~\cite{weiss2020deepracing}, demonstrate the potential of learning for high-speed autonomy, but they do not provide an explicit dynamics model suitable for nonlinear model predictive control (NMPC). 
Other learning-based racing and autonomy frameworks, including Gaussian-process reinforcement learning with collision avoidance~\cite{lu2024gpskrl} and digital-twin contextual reinforcement learning~\cite{ju2023digitaltwin}, further highlight the value of adaptation, but often retain limited physical transparency in the learned representation.

Hybrid vehicle dynamics models aim to combine the interpretability of analytical models with the flexibility of learning-based correction or adaptation. 
The Deep Dynamics Model (\ddm{})~\cite{chrosniak2024deep} uses a physics-constrained neural network to estimate vehicle dynamics parameters within a single-track model. 
This design improves interpretability compared with purely neural predictors, but its learned representation is not explicitly organized around long-memory state-space adaptation and short-term recurrent transient response. 
The Fine-Tuning Hybrid Dynamics model (\fthd{})~\cite{fang2025fine} augments a physics model with a residual physics-informed neural network and incorporates filtering for robustness under limited data. 
While effective in open-loop prediction settings, residual-heavy hybrid formulations can remain sensitive to how the learned correction behaves when recursively evaluated inside a controller. 
Other approaches incorporate Gaussian-process residuals into model predictive control~\cite{li2025learning}, use physics-informed neural networks for slip-related dynamics~\cite{kolluri2025physics}, or employ Koopman-based representations with adaptive constraints~\cite{zhang2024koopman}. 
These methods show that embedding physical structure into learned dynamics can improve generalization and control relevance, but many still treat temporal representation, physical parameter adaptation, and residual correction as separate or loosely coupled modules.

Vehicle dynamics are inherently history dependent: the current response depends not only on the instantaneous state and command input, but also on recent actuator feedback, command history, tire-force evolution, and varying operating conditions. 
Recurrent neural networks \cite{kebbati2025rnn}, including GRU- and LSTM-based models \cite{zhang2024vehicle}, provide a natural mechanism for capturing such temporal dependencies and have been widely adopted for sequence prediction tasks. 
However, recurrent models can become computationally expensive or difficult to train over longer histories, and their hidden states are not directly tied to interpretable physical quantities. 
State-space sequence models provide an alternative representation by maintaining an evolving latent state with efficient temporal updates. 
Linear state-space layers have been shown to combine recurrent, convolutional, and continuous-time modeling perspectives for sequence learning~\cite{gu2021lss}. 
More recent work has shown that state-space models can serve as accurate and efficient neural operators for dynamical systems, making them promising for compact modeling of temporal evolution~\cite{hu2025mamba}. 
These characteristics are particularly attractive for vehicle dynamics modeling, where both long-term operating context and short-term transient behavior must be represented accurately while maintaining computational efficiency for real-time deployment.

Motivated by these challenges, we propose \method{}, a physics-aware dynamics modeling framework that integrates temporal memory, bounded physics-guided adaptation, and residual correction within a unified prediction architecture. The specific contributions of this work are as follows:
\begin{enumerate}[noitemsep]
    \item We propose a physics-aware state-space memory framework (\method{}) that combines structured vehicle dynamics with state-space sequence modeling and recurrent encoding to capture both long-term operating context and short-term actuator-induced transients.
    \item We introduce a bounded physics-guided adaptation and residual refinement mechanism that adjusts physically meaningful vehicle dynamics parameters (including tire-force, slip-shift, force-shift, and inertial coefficients) based on learned temporal features, while a gated residual correction compensates for remaining model mismatch and unmodeled effects without sacrificing physical interpretability.
    \item We provide a comprehensive evaluation of \method{} using simulated small-scale racing data and real-world autonomous Indy racecar data, including open-loop prediction, component-wise ablation studies, and closed-loop nonlinear model predictive control (NMPC) simulations to assess prediction accuracy, computational efficiency, and controller feasibility.
\end{enumerate}
 
The remainder of the paper is organized as follows. 
Sec.~\ref{sec:mathmodel} presents the physics-guided vehicle model, including slip-angle dynamics and lateral and longitudinal load-transfer effects. 
Sec.~\ref{sec:dynssm_framework} describes the proposed \method{} framework. 
Sec.~\ref{sec:datagenerration} details the data-generation process and experimental datasets. 
Sec.~\ref{sec:Results} presents the open-loop and closed-loop evaluation results. Finally, 
Sec.~\ref{sec:conclusion} concludes the paper and discusses future research directions.

\section{Mathematical Model}
\label{sec:mathmodel}

\begin{table}[htbp]
\caption{State, Control, Learning, and Dynamics Variables Used in the \method{}}
\label{tab:var_categories}
\centering
\scriptsize

\begin{tblr}{
row{1-2} = {font=\bfseries\scriptsize},
row{2} = {bg=brown8,fg=black,font=\bfseries\scriptsize},
row{3-7} = {font=\scriptsize},
colspec = {X[0.25,c,m] X[0.52,c,m] X[0.13,c,m]},
colsep = .8mm,
rowsep = .2mm,
vlines = {0.8pt, black},
hlines = {0.8pt, black},
hline{4-7} = {dashed,fg=gray},
}
Symbol & Description & Units \\
\SetCell[c=3]{c} Pose Variables for Rollout/Evaluation & & \\
$x$, $y$ & Global position in world frame & \si{\meter} \\
$\theta$ & Yaw angle/heading & \si{\radian} \\
$v_x$ & Longitudinal velocity in body frame & \si{\meter\per\second} \\
$v_y$ & Lateral velocity in body frame & \si{\meter\per\second} \\
$\omega$ & Yaw rate & \si{\radian\per\second} \\
\end{tblr}

\vspace{0.5mm}

\begin{tblr}{
row{1} = {bg=azure8,fg=black,font=\bfseries\scriptsize},
row{2-7} = {font=\scriptsize},
colspec = {X[0.25,c,m] X[0.52,c,m] X[0.13,c,m]},
colsep = .8mm,
rowsep = .2mm,
vlines = {0.8pt, black},
hlines = {0.8pt, black},
hline{3-7} = {dashed,fg=gray},
}
\SetCell[c=3]{c} Actuator Feedback and Command Inputs & & \\
$r_{\mathrm{th,fb}}$ & Throttle feedback & \si{\percent} \\
$\delta_{\mathrm{fb}}$ & Steering feedback & \si{\radian} \\
$r_{\mathrm{cmd}}$ & Throttle command & \si{\percent} \\
$\delta_{\mathrm{cmd}}$ & Steering command & \si{\radian} \\
$r_{\mathrm{app},t}$ & Applied throttle input & \si{\percent} \\
$\delta_t$ & Applied steering input & \si{\radian} \\
\end{tblr}

\vspace{0.5mm}

\begin{tblr}{
row{1} = {bg=gray8,fg=black,font=\bfseries\scriptsize},
row{2-24} = {font=\scriptsize},
colspec = {X[0.25,c,m] X[0.52,c,m] X[0.13,c,m]},
colsep = .8mm,
rowsep = .2mm,
vlines = {0.8pt, black},
hlines = {0.8pt, black},
hline{3-23} = {dashed,fg=gray},
}
\SetCell[c=3]{c} Dynamics and Vehicle Parameters & & \\
$l_f$, $l_r$ & Distance from CG to front/rear axle & \si{\meter} \\
$h_{\mathrm{cg}}$ & Height of center of gravity & \si{\meter} \\
$m$ & Vehicle mass & \si{\kilogram} \\
$I_z$ & Yaw moment of inertia & \si{\kilogram\meter\squared} \\
$g$ & Gravitational acceleration & \si{\meter\per\second\squared} \\
$\alpha_f$, $\alpha_r$ & Front/rear slip angles & \si{\radian} \\
$\psi_f$, $\psi_r$ & Front/rear tire shape functions & \si{\radian} \\
$F_{rx}$ & Rear longitudinal tire force & \si{\newton} \\
$F_{fy}$, $F_{ry}$ & Front/rear lateral tire forces & \si{\newton} \\
$F_{zf}$, $F_{zr}$ & Front/rear normal tire loads & \si{\newton} \\
$F_{z0}$ & Nominal static normal load & \si{\newton} \\
$C_{m1}$, $C_{m2}$ & Drive-force coefficients & -- \\
$C_{r0}$, $C_{r2}$ & Rolling resistance coefficients & -- \\
$B_f$, $C_f$, $D_f$, $E_f$ & Front tire parameters & -- \\
$B_r$, $C_r$, $D_r$, $E_r$ & Rear tire parameters & -- \\
$S_{hf}$, $S_{hr}$ & Front/rear slip-angle shifts & \si{\radian} \\
$S_{vf}$, $S_{vr}$ & Front/rear lateral-force shifts & \si{\newton} \\
$a_x$ & Longitudinal acceleration & \si{\meter\per\second\squared} \\
$a_y$ & Lateral acceleration & \si{\meter\per\second\squared} \\
$\dot{\omega}$ & Yaw acceleration & \si{\radian\per\second\squared} \\
$f_{\mathrm{dyn}}(\cdot)$ & Physics-guided dynamics layer & -- \\
$r_{\psi}(\cdot)$ & Residual correction network & -- \\
\end{tblr}
\end{table}

\subsection{Dynamic Single-Track Model} \label{subsec:physics_aware_stm}
A single-track vehicle dynamics model is defined as \cite{jazar2025vehicle}, 
\begin{eqnarray}
    \dot{v}_{x,t} &=& \frac{1}{m} (F_{rx,t}  -  F_{fy,t}\sin\delta_t ) + v_{y,t}\omega_t , \label{eq:vx_dot}   \\
    \dot{v}_{y,t} &=& \frac{1}{m} (F_{ry,t}  +  F_{fy,t}\cos\delta_t ) - v_{x,t}\omega_t , \label{eq:vy_dot} \\
    \dot{\omega}_{t} &=& \frac{ l_f F_{fy,t}\cos\delta_t   -   l_r F_{ry,t} }{I_z} . \label{eq:w_dot}
\end{eqnarray}
The continuous-time dynamic state is defined as $\mathbf{s}_t=[v_{x,t},v_{y,t},\omega_t]^\top$, where $v_{x,t}$ and $v_{y,t}$
are the longitudinal and lateral body-frame velocities, and $\omega_t$ is the
yaw rate.
Table~\ref{tab:var_categories} details all the parameters described in Eq. (\ref{eq:vx_dot})-(\ref{eq:w_dot}).

\begin{figure}[ht]
    \centering
    \includegraphics[width=1\linewidth]{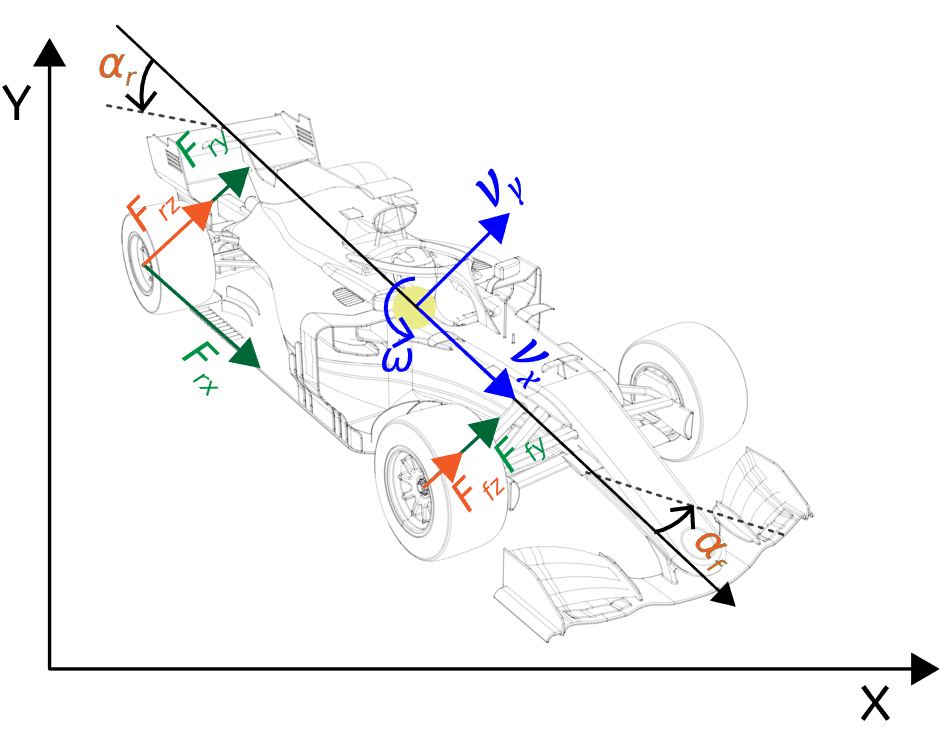}
    \caption{Single-track vehicle model used in \method{}. State information shown in blue $[\vx, \vy, \w]^\top$, along with slip angles $\af{}$ and $\ar{}$, normal loads \Ffz{} and \Frz{} computed with longitudinal load transfer, and tire forces \Frx{}, \Ffy{}, \Fry{}.}
    \label{fig:singletrack}
\end{figure} 
\subsection{Enhanced Physics-based Vehicle Dynamics} \label{sec:physics_guided_dynamics}
Although the single-track structure provides an interpretable analytical
backbone, accurate racing dynamics require additional effects beyond a fixed
nominal model.
\method{} incorporates these additional effects through a physics-guided dynamics layer that includes a 1) front/rear slip angle; 2) load transfer effects and 3) 
load-dependent Pacejka-style lateral tire forces. 
\subsubsection{Slip Angle} \label{subsec:slip_tire_force}
The slip angles $\af{}$ and $\ar{}$ quantify the angular difference between the orientation of the tire and its actual direction of travel, which is crucial for lateral tire force generation. They are computed as:
\begin{eqnarray}
    \alpha_f &=& \delta - \tan^{-1}\bigg( \frac{l_f \omega + v_y}{|v_x|}  \bigg)  +  S_{hf},  \label{eq:front_slip} \\
    \alpha_r &=& \tan^{-1}\bigg( \frac{l_r \omega - v_y}{|v_x|} \bigg) + S_{hr}. \label{eq:rear_slip}
\end{eqnarray}
\subsubsection{Load Transfer Effects} \label{subsec:longitudinal_lateral_dynamics}
During acceleration and braking, the vertical load on each axle shifts due to inertia and suspension geometry. We model longitudinal load transfer in the bicycle model via static equilibrium of vertical forces and pitch moments about the CG. 
The longitudinal tire force is modeled as
\begin{equation}
    F_{rx}
    =
    \left(C_{m1}-C_{m2}v_x\right)r_{\mathrm{th}}
    -
    C_{r0}
    -
    C_{r2}v_x^2 .
\label{eq:longitudinal_force}
\end{equation}
For the full-scale IAC model, the longitudinal force–induced acceleration along with front and rear normal tire loads are modeled as,
\begin{eqnarray}
    a_x^{\mathrm{long}} &=& \frac{F_{rx}}{m}, \\
    F_{zf} & = & \frac{m g l_r - h_{\mathrm{cg}} m a_x^{\mathrm{long}}}{ l_f+l_r}, \\
    F_{zr} & = & \frac{m g l_f + h_{\mathrm{cg}} m a_x^{\mathrm{long}}}{ l_f+l_r} . \label{eq:dynamic_load_transfer}
\end{eqnarray}
\subsubsection{Lateral Tire Forces}
The lateral tire forces are modeled using a load-dependent Pacejka formulation with curvature as 
shape functions as:
\begin{eqnarray}
    F_{fy} &=& \left[S_{vf} + D_f\sin(C_f\psi_f) \right] \frac{F_{zf}}{F_{z0}}, \\
    F_{ry} &=& \left[S_{vr} + D_r\sin(C_r\psi_r) \right] \frac{F_{zr}}{F_{z0}} 
\label{eq:tire_forces}
\end{eqnarray}
where $\psi_f, \psi_r$ are the front and rear tire shape functions denoted as: 
\begin{eqnarray}
   \psi_f &=& \tan^{-1} \left[B_f\alpha_f - E_f \left( B_f\alpha_f - \tan^{-1}(B_f\alpha_f) \right)
    \right], \\
    \psi_r &= & \tan^{-1} \left[B_r\alpha_r - E_r \left( B_r\alpha_r - \tan^{-1}(B_r\alpha_r)    \right) \right]. \label{eq:tire_shape_functions}
\end{eqnarray}
Here, $F_{z0}$ denotes the nominal vertical load, while $F_{zf}/F_{z0}$ and $F_{zr}/F_{z0}$ introduce load sensitivity into the front and rear lateral force generation. 
The overall single-track formulation, including slip angles, normal loads, tire forces, and state vectors are illustrated in \figref{fig:singletrack}.

\section{Proposed DynSSM Framework}
\label{sec:dynssm_framework}

\subsection{Architecture Overview}
\label{subsec:architecture_overview}

\begin{figure*}[t]
\noindent
\makebox[\textwidth][c]{%
 \begin{adjustbox}{max width=\textwidth,max height=\textheight}
 
\begin{tikzpicture}[
    font=\small,
    >=Stealth,
    line/.style={->,line width=0.75pt},
    blk/.style={
        draw=black,
        rounded corners=4pt,
        line width=0.75pt,
        fill=white,
        align=center
    },
    blueblk/.style={
        draw=memblue,
        rounded corners=5pt,
        line width=0.9pt,
        fill=white,
        align=center
    },
    orangeblk/.style={
        draw=physorange,
        rounded corners=5pt,
        line width=0.9pt,
        fill=white,
        align=center
    },
    purpleblk/.style={
        draw=respurple,
        rounded corners=5pt,
        line width=0.9pt,
        fill=white,
        align=center
    }
    ]
    \path[use as bounding box] (-0.5,-6.7) rectangle (17.8,6.0);
			\node[blk, minimum width=2.6cm, minimum height=3.3cm] (input) at (3,4) {
				{\bfseries Input History Window $H_t$}\\[3mm]
				$H_t=[x_{t-H+1},u_{t-H+1} \ldots,x_t,u_t]$\\[3mm]
				$x_t=[v_x,v_y,\omega,r_{fb},\delta_{fb}]$\\[3mm]
				$u_t=[r_{cmd},\delta_{cmd}]$
			};
			
			\node[
			blueblk,
			minimum width=8cm,
			minimum height=4.5cm,
			fill=softblue
			] (memorybox) at (11,4.05) {};
			
			\node[text=memblue,font=\bfseries\Large] at ($(memorybox.north)+(0,-0.35)$)
			{Neural Memory Branch};
			
			\node[text=memblue,font=\itshape\large] at ($(memorybox.north)+(0,-0.78)$)
			{SSM + GRU temporal encoding};
			
			\node[blueblk, minimum width=3.1cm, minimum height=1.0cm] (ssm) at (9,4.4) {
				{\bfseries SSM Encoder}\\[-1mm]
				{\itshape long-memory}\\[-1mm]
				{\itshape operating context}
			};
			
			\node[blueblk, minimum width=3.1cm, minimum height=1.0cm] (gru) at (9,3.1) {
				{\bfseries GRU Encoder}\\[-1mm]
				{\itshape short-term}\\[-1mm]
				{\itshape transient response}
			};
			
			\node[blueblk, minimum width=3.45cm, minimum height=0.55cm] (latest) at (9.3,2.18) {
				latest normalized input $\bar h_t$
			};
			
			\node[blueblk, minimum width=1.7cm, minimum height=3.05cm] (fusion) at (13.75,3.44) {
				{\bfseries Feature}\\
				{\bfseries Fusion}
			};
			
			\draw[line] (ssm.east) -- ++(2.3,0);
			\draw[line] (gru.east) -- ++(2.3,0);
			\draw[line] (latest.east) -- ++(1.8,0);
			
			\draw[line] (input.east) -- ++(0,0) |- (memorybox.west);
    
    \node[
    orangeblk,
    minimum width=15.0cm,
    minimum height=4.5cm,
    fill=softorange
    ] (physbox) at (10.0,-1) {};
    
    \node[text=physorange,font=\bfseries\Large] at ($(physbox.north)+(0,-0.32)$)
    {Structured Physics Branch};
    
    \node[text=physorange,font=\itshape\large] at ($(physbox.north)+(0,-0.74)$)
    {Physics-guided single-track dynamics model};
    
    \node[orangeblk, minimum width=4cm, minimum height=2.25cm, align=left] (adapt) at (5.2,-1.3) {
        \begin{minipage}{4.3cm}
            \centering
            {\scriptsize \textbf{Bounded Parameter Adaptation}}\\ 
            $\theta_t=\mathrm{clip}(\theta_0+\Delta\theta_t,\theta_{\min},\theta_{\max})$\\[-0.5mm]
            \tikz{\draw[physorange,dashed,line width=0.6pt] (0,0) -- (2.8,0);}\\ 
            \raggedright
            {\itshape Adapted parameters (examples):}\\ 
            {\itshape $\bullet$ tire parameters}\\ 
            {\itshape $\bullet$ slip/force shifts}\\ 
            {\itshape $\bullet$ drive/resistance terms}\\ 
            {\itshape $\bullet$ yaw inertia}
        \end{minipage}
    };

 			\draw[line, ->]
 			(latest.south)
 			-- ++(0,-0.3)
 			-- ++(-3.5,0) -- ++(0,-0.2)
 			to[out=180,in=90] ++(-0.2,-0.2)
 			to[out=-90,in=180] ++(0.2,-0.2)
 			-- ++(0,-0.75);

\node[
    anchor=west,
    align=left,
    font=\scriptsize
] at ($(fusion.east)+(0.55,0.32)$) {
    fused latent\\[-1mm]
    memory $\mathbf{z}_t$
};

\draw[line, ->] (fusion.east) to ++(1,0) to ++(0,-2.2);

    \node[orangeblk, minimum width=9.6cm, minimum height=3.3cm] (dynbox) at (12.5,-1.3) {};
    \node[font=\bfseries] at ($(dynbox.north)+(0,-0.28)$)
    {Physics-Guided Single-Track Dynamics};

\draw[line, ->] (input.south) to ++(0,-1.1);

%
%

\node[
    blk,
    minimum width=1cm,
    minimum height=2cm,
    align=center,
    inner sep=2pt
] (m1) at (8.5,-1.5) {
    {\scriptsize applied}\\[-1mm]
    {\scriptsize inputs}\\[1mm]
    \includegraphics[width=1.2cm]{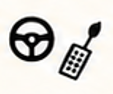}\\[1mm]
    {\tiny $(r_{\mathrm{cmd}},\delta_{\mathrm{cmd}})$}\\[-1mm]
};

\node[
    blk,
    minimum width=1cm,
    minimum height=2cm,
    align=center,
    inner sep=2pt,
    right=0.2cm of m1
    ] (m2)  {
    {\scriptsize slip}\\[-1mm]
    {\scriptsize angles}\\[1mm]
    \includegraphics[width=1.0cm]{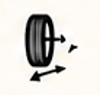}\\[1mm]
    {\tiny $(\alpha_f,\alpha_r)$}\\[-1mm]
};

%

\node[
    blk,
    minimum width=1cm,
    minimum height=2cm,
    align=center,
    inner sep=2pt,
    right=0.2cm of m2
    ] (m3)  {
    {\scriptsize load}\\[-1mm]
    {\scriptsize transfer}\\[1mm]
    \includegraphics[width=1.4cm]{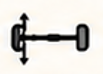}\\[1mm]
    $\frac{1}{\Delta F_z}$\\[-1mm]
};

%

\node[
    blk,
    minimum width=1cm,
    minimum height=2cm,
    align=center,
    inner sep=2pt,
    right=0.2cm of m3
    ] (m4)  {
    {\scriptsize Tire}\\[-1mm]
    {\scriptsize forces}\\[1mm]
    \includegraphics[width=1.3cm]{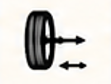}\\[1mm]
    {\tiny $(F_{yf},F_{yr})$}\\[-1mm]
};


\node[
    blk,
    minimum width=1cm,
    minimum height=2cm,
    align=center,
    inner sep=2pt,
    right=0.2cm of m4
    ] (m5)  {
    {\scriptsize Accelerations}\\[1mm]
    \includegraphics[width=1.3cm]{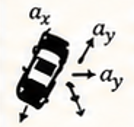}\\[1mm]
    {\tiny $(a_{x},a_{y},\hat{\omega})$}\\[-1mm]
};

\node[
    blk,
    minimum width=1cm,
    minimum height=2cm,
    align=center,
    inner sep=2pt,
    right=0.2cm of m5
    ] (m6)  {
    {\scriptsize State}\\[-1mm]
    {\scriptsize propagation}\\[1mm]
    \includegraphics[width=0.95cm]{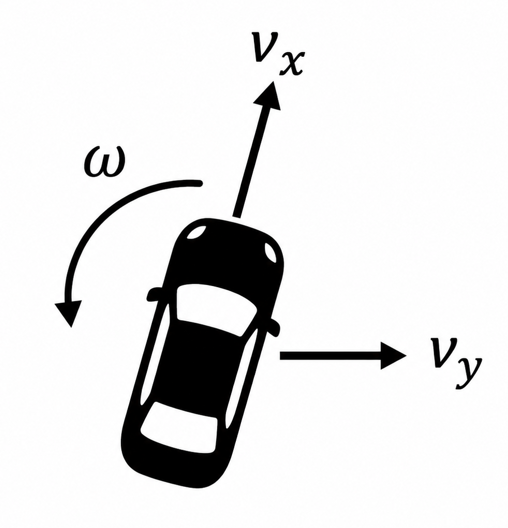}\\[1mm]
    {\tiny $(\dot v_x,\dot v_y,\dot\omega)$}\\[-1mm]
};

%
%
%
%
%
%
    \node[purpleblk, minimum width=4cm, minimum height=2.45cm, fill=softpurple] (resid) at (8.5,-5) {};

\draw[line, -]
([xshift=-3.75cm]physbox.south)
-- ([xshift=-3.75cm,yshift=-2.25cm]physbox.south |- resid.north);

    \node[text=respurple,font=\bfseries\large,centered] at ($(resid.north)+(0,-0.38)$)
    {Gated Residual};

    \node[text=respurple,font=\bfseries\large] at ($(resid.north)+(0,-0.78)$)
    {Correction};
%
    \foreach \i/\y in {1/-5,2/-5.5,3/-6} {
        \node[circle,draw=respurple,inner sep=1.5pt] (ri\i) at (7,\y) {};
    }
    \foreach \i/\y in {1/-5,2/-5.5,3/-6} {
        \node[circle,draw=respurple,inner sep=1.5pt] (rh\i) at (8,\y) {};
    }
    \foreach \i in {1,2,3}{
        \foreach \j in {1,2,3}{
            \draw[respurple,line width=0.3pt] (ri\i) -- (rh\j);
        }
    }
    \draw[line,draw=respurple] (8.1,-5.5) -- (8.7,-5.5);
    \node[circle,font=\large,draw=respurple,minimum size=0.75cm] at (9.3,-5.5) {$ \textcolor{black}{\text{\faLock}}$};
%
    \coordinate (feedx) at (6.15,-4);
    \node[anchor=east] at (1.05,-3.8) {$c_t$};
    \node[anchor=west,font=\itshape] at (1.05,-3.8) {(physics-conditioned feature)};
    \node[anchor=east] at (1.05,-4.3) {$\tilde{\boldsymbol{\theta}}_t$};
    \node[anchor=west,font=\itshape] at (1.05,-4.3) {(normalized parameter deviation)};
    \node[anchor=east] at (1.15,-4.8) {$h_t$};
    \node[anchor=west,font=\itshape] at (1.05,-4.8) {(latest raw state-action input)};
    \node[anchor=east] at (1.05,-5.3) {$\hat s^{\rm phy}_{t+1}$};
    \node[anchor=west,font=\itshape] at (1.05,-5.3) {(physics prediction)};
    \node[anchor=east] at (1.05,-5.8) {$\Delta\hat s^{\rm phy}_{t}$};
    \node[anchor=west,font=\itshape] at (1.05,-5.8) {(prediction residual proxy)};
    
    \foreach \y in {-4,-4.5,-5,-5.5,-6}{
        \draw[line] (3.5,\y) -- (resid.west |- 0,\y);
    }
    
    \draw[line] ($(physbox.south)+(2,0)$) -- ++(0,-1.4);
    \node[anchor=west,align=center] at (12,-3.8) {physics prediction $\hat s^{\rm phy}_{t+1}$};

    \draw[line] ($(resid.east)+(0,-0.0)$) -- ++(1.1,0);
    \node[anchor=west, align=center] at (10.6,-5.45) {gated\\residual};
    \node[circle,draw=black,line width=0.9pt,minimum size=0.72cm] (sum) at (12,-5) {\Large $+$};
    
			\node[blk, minimum width=4cm, minimum height=1.1cm] (out) at (15.5,-5) {
				{\bfseries Final Next-State}\\ 
				{\bfseries Prediction}\\[2mm]
				$\hat s_{t+1}= [\hat v_{x,t+1},\hat v_{y,t+1},\hat\omega_{t+1}]^T$
			};
	\node[text=respurple,font=\itshape,align=center] at ($(out.south)+(-0.25,-0.65)$)
	{Used for open-loop prediction \\ and closed-loop control};
    \draw[line] ($(sum.east)+(0,-0.0)$) -- ++(1.1,0);
    \end{tikzpicture}
    \end{adjustbox} 
    }
 \caption{Overview of the proposed DynSSM architecture. The neural memory branch uses parallel SSM and GRU encoders to extract long- and short-term features from state-action history. The structured physics branch produces the primary physics-based next-state prediction in conjunction with the bounded parameter adaptation. A gated residual correction then refines this physics prediction to obtain the final next-state estimate used for open-loop prediction and closed-loop NMPC simulation.}
 
    \label{fig:nn_archi}
\end{figure*}	

Fig.~\ref{fig:nn_archi} illustrates the proposed \method{} framework.
The model is organized as a hierarchical prediction pipeline that couples
neural memory branch (with SSM and GRU temporal encoding), structured physics branch (with physics-guided single track dynamics model along with bounded parameter adaptation) and finally with a gated residual correction. 
Along with the dynamic state $\mathbf{s}_t$, \method{} receives a history of  actuator-feedback, and command inputs that are defined as:
%
%
\begin{subequations}
    \begin{align}
        \mathbf{x}_t =& [v_{x,t},v_{y,t},\omega_t,r_{\mathrm{fb},t},\delta_{\mathrm{fb},t}]^\top, \label{eq:model_state} \\
        \mathbf{u}_t =& [r_{\mathrm{cmd},t},\delta_{\mathrm{cmd},t}]^\top. \label{eq:model_input}
    \end{align}
\end{subequations} 
The applied throttle and steering used by the analytical dynamics layer are
constructed from the latest feedback and command signals: 
\begin{equation}
    r_{\mathrm{app},t}
    =
    r_{\mathrm{fb},t}
    +
    r_{\mathrm{cmd},t},
    \qquad
    \delta_t
    =
    \delta_{\mathrm{fb},t}
    +
    \delta_{\mathrm{cmd},t}.
\label{eq:problem_applied_inputs}
\end{equation}
No explicit hand-tuned actuator-rate limiter is imposed inside the dynamics
rollout.
The variables used throughout the \method{} architecture are summarized in Table~\ref{tab:nn_params}.
\begin{table}[htbp]
\caption{Parameters in \method{} Architecture}
\label{tab:nn_params}
\centering
\scriptsize

\begin{tblr}{
row{1} = {bg=green8,fg=black,font=\bfseries\scriptsize},
row{2-18} = {font=\scriptsize},
colspec = {X[0.25,c,m] X[0.52,c,m] X[0.13,c,m]},
colsep = .8mm,
rowsep = .2mm,
vlines = {0.8pt, black},
hlines = {0.8pt, black},
hline{3-17} = {dashed,fg=gray},
}
\SetCell[c=3]{c} Learning and Adaptation Variables & & \\
$\mathbf{H}_t$ & State-action history window & -- \\
$\mathbf{x}_t$ & Raw input state & -- \\
$\mathbf{u}_t$ & Command input vector & -- \\
$\mathbf{s}_t$ & Predicted dynamic state & -- \\
$\bar{\mathbf{h}}_t$ & Latest normalized input & -- \\
$\mathbf{z}_t$ & Fused temporal feature & -- \\
$\mathbf{c}_t$ & Physics-conditioned feature & -- \\
$\boldsymbol{\theta}_t$ & Adapted parameter vector & -- \\
$\boldsymbol{\theta}_0$ & Nominal parameter vector & -- \\
$\boldsymbol{\theta}_{\min}$, $\boldsymbol{\theta}_{\max}$ & Parameter bounds & -- \\
$\tilde{\boldsymbol{\theta}}_t$ & Normalized parameter deviation & -- \\
$\mathbf{s}_y$ & Target normalization scale & -- \\
$\mathbf{s}_{\Delta}$ & State-increment scale & -- \\
$\mathbf{g}_r$ & Residual gate & -- \\
$\mathbf{r}^{\mathrm{corr}}_t$ & Normalized residual-network output & -- \\
$T_s$ & Sampling time & \si{\second} \\
\end{tblr}

\end{table}

\subsection{Temporal Encoder and Parameter Adaptation}
\label{subsec:temporal_encoder_param_adapt}

Given a history window
\begin{equation}
\mathbf{H}_t =
\left[
\mathbf{x}_{t-H+1},\mathbf{u}_{t-H+1},
\ldots,
\mathbf{x}_{t},\mathbf{u}_{t}
\right],
\end{equation}
where $\mathbf{x}_t, \mathbf{u}_t$ defined in Eqs.(\ref{eq:model_state})-(\ref{eq:model_input}), the normalized state-action sequence is passed through two complementary temporal encoders.
The SSM branch captures slowly varying history-dependent behavior, while the  
GRU branch captures short-term transient responses. These two are fused with the latest normalized input $\bar{\mathbf{h}}_t$ to obtain a compact latent representation,
%
\begin{equation}
\mathbf{z}_t =
\phi_{\mathrm{fuse}}
\left(
\phi_{\mathrm{SSM}}(\mathbf{H}_t),
\phi_{\mathrm{GRU}}(\mathbf{H}_t),
\bar{\mathbf{h}}_t
\right),
\end{equation}
%
%
The latent representation is then passed to a physics-conditioning network that
predicts bounded deviations from nominal vehicle-dynamics parameters:
\begin{subequations}
\begin{alignat}{2}
&\Delta\boldsymbol{\theta}_t
\qquad
&&= \alpha_{\theta}
\tanh\left(
g_{\theta}\left(
\mathbf{z}_t,\bar{\mathbf{h}}_t
\right)
\right)
\odot
\Delta\boldsymbol{\theta}_{\mathrm{range}}, \\
&\Delta\boldsymbol{\theta}_{\mathrm{range}}
&&=
\boldsymbol{\theta}_{\max}
-
\boldsymbol{\theta}_{\min}, \label{eq:del_theta}\\
&\boldsymbol{\theta}_t
&&=
\mathrm{clip}\left(
\boldsymbol{\theta}_0+\Delta\boldsymbol{\theta}_t,
\boldsymbol{\theta}_{\min},
\boldsymbol{\theta}_{\max}
\right),
\label{eq:bounded_param_adapt}
\end{alignat}
\end{subequations}

where $\boldsymbol{\theta}_0$ denotes the nominal parameter vector, $\Delta\boldsymbol{\theta}_{\mathrm{range}}$ defines the admissible parameter range, and $\alpha_{\theta}$ controls the adaptation magnitude. This bounded formulation keeps the learned parameterization physically admissible while allowing the model to adapt to local operating conditions.
\subsection{Physics-Informed Learning Objective}
\label{subsec:physics_informed_objective}

The model is trained using one-step supervised prediction over the dynamic
state vector $\mathbf{s}_t=[v_x,v_y,\omega]^\top$. Let
$\hat{\mathbf{s}}_{i,t+1}$ denote the predicted dynamic state for sample $i$,
and let $\mathbf{s}^{\mathrm{obs}}_{i,t+1}$ denote the corresponding observed
target. The primary prediction loss is a target-normalized mean-squared error:
\begin{equation}
\mathcal{L}_{\mathrm{pred}}
=
\frac{1}{N d_s}
\sum_{i=1}^{N}
\left\|
\frac{
\hat{\mathbf{s}}_{i,t+1}
-
\mathbf{s}^{\mathrm{obs}}_{i,t+1}
}
{\mathbf{s}_y}
\right\|_2^2 ,
\label{eq:prediction_loss}
\end{equation}
where $d_s$ is the dimension of the predicted dynamic state and
$\mathbf{s}_y$ is the state-wise target normalization scale. This normalization
prevents high-magnitude states from dominating the training objective and makes
the errors in longitudinal velocity, lateral velocity, and yaw rate comparable
during optimization.

To preserve physical plausibility, the prediction loss is augmented with a
parameter regularization term that discourages excessive deviation from the
nominal dynamics parameters:
\begin{equation}
\mathcal{L}_{\theta}
=
\lambda_{\theta}
\frac{1}{d_{\theta}}
\left\|
\tilde{\boldsymbol{\theta}}_t
\right\|_2^2,
\qquad
\tilde{\boldsymbol{\theta}}_t
=
\frac{
\boldsymbol{\theta}_t-\boldsymbol{\theta}_0
}
{
\Delta\boldsymbol{\theta}_{\mathrm{range}}
},
\label{eq:param_reg_loss}
\end{equation}
where $d_{\theta}$ is the number of adapted physical parameters and
$\Delta\boldsymbol{\theta}_{\mathrm{range}}$ is defined in Eq. (\ref{eq:del_theta}).
The normalized parameter deviation $\tilde{\boldsymbol{\theta}}_t$ measures how
far the adapted parameter vector moves from the nominal values relative to the
admissible parameter span.

For configurations that explicitly constrain the residual pathway, two
additional regularizers are used. The residual penalty discourages large gated
corrections after the physics prediction, while the physics-anchor term
penalizes the normalized gap between the final prediction and the physics-only
prediction. The actual correction added after the physics rollout is
\begin{equation}
\Delta\hat{\mathbf{s}}^{\mathrm{corr}}_t
=
\hat{\mathbf{s}}_{t+1}
-
\hat{\mathbf{s}}^{\mathrm{phy}}_{t+1}
\label{eq:actual_residual_correction}
\end{equation}
The residual and physics-anchor losses are then written as

\begin{eqnarray}
\mathcal{L}_{\mathrm{corr}}
&= &
\lambda_{\mathrm{corr}}
\frac{1}{d_s}
\left\|
\mathbf{r}^{\mathrm{corr}}_t
\right\|_2^2 , \label{eq:residual_loss} \\
%
\mathcal{L}_{\mathrm{phy}}
&= &
\lambda_{\mathrm{phy}}
\frac{1}{d_s}
\left\|
\frac{
\hat{\mathbf{s}}_{t+1}
-
\hat{\mathbf{s}}^{\mathrm{phy}}_{t+1}
}
{\mathbf{s}_{\Delta}}
\right\|_2^2 .
\label{eq:physics_anchor_loss}
\end{eqnarray}

%
where $\mathbf{r}^{\mathrm{corr}}_t$ is the normalized residual-network output as defined in Eq.~\eqref{eq:residual_output}, $\lambda_{phy}$ controls the residual magnitude, and $\mathbf{s}_{\Delta}$ is the state-wise delta scale. 
The residual loss penalizes large residual-network outputs, whereas the physics-loss penalizes the normalized deviation between the final and physics-only prediction. The vector $\mathbf{s}_{\Delta}$ denotes the state-wise delta scale used to normalize state increments.
The complete objective can be written as
\begin{equation}
\mathcal{L}
=
\mathcal{L}_{\mathrm{pred}}
+
\mathcal{L}_{\theta}
+
\mathbb{I}_{\mathrm{corr}}\mathcal{L}_{\mathrm{corr}}
+
\mathbb{I}_{\mathrm{phy}}\mathcal{L}_{\mathrm{phy}},
\label{eq:total_loss}
\end{equation}
where $\mathbb{I}_{\mathrm{corr}}$ and $\mathbb{I}_{\mathrm{phy}}$ indicate
whether the residual and physics-anchor regularizers are enabled for a given experimental configuration.

The full-scale IAC and ORCA-scale configurations differ primarily in vehicle scale, operating regime, and available data sources. The IAC configuration corresponds to full-scale autonomous Indy racecars operating at high speeds, where the data are collected from real AV-21 vehicles equipped with multi-modal sensing systems \cite{kulkarni2023racecar}. In contrast, the ORCA-scale configuration follows the small-scale autonomous racing setting commonly used for model-based racing control studies, where learning-based dynamics correction can be evaluated under controlled and repeatable conditions \cite{jain2020bayesrace}.

In the full-scale IAC configuration, the objective uses target-normalized prediction loss with parameter regularization. In the ORCA-scale configuration, residual and physics-anchor regularization are also enabled to further limit unstructured corrections. Together with the bounded parameterization, this objective encourages the model to rely on the adaptive physics-guided dynamics layer as the primary prediction mechanism while allowing the residual branch to correct only the remaining local mismatch.

\subsection{Gated Residual Correction} \label{subsec:gated_residual}
The adapted dynamics layer produces the primary physics-based prediction for
the dynamic state $\mathbf{s}_t=[v_x,v_y,\omega]^\top$:
\begin{equation}
\hat{\mathbf{s}}^{\mathrm{phy}}_{t+1}
=
f_{\mathrm{dyn}}(\mathbf{x}_t,\mathbf{u}_t;\boldsymbol{\theta}_t),
\label{eq:physics_prediction}
\end{equation}
where $\mathbf{x}_t$ and $\mathbf{u}_t$ denote the most recent raw state and
control inputs used by the analytical dynamics layer, and
$\boldsymbol{\theta}_t$ is the adapted parameter vector. This physics-based
prediction preserves the structured vehicle-dynamics computation, but real-world
effects such as actuator delay, tire-saturation mismatch, local surface
variation, and measurement uncertainty may still introduce residual model error.
To compensate for these remaining discrepancies, \method{} applies a gated
residual correction after the physics rollout. The physics-predicted state
increment is first defined as
\begin{equation}
\Delta\hat{\mathbf{s}}^{\mathrm{phy}}_t
=
\hat{\mathbf{s}}^{\mathrm{phy}}_{t+1}
-
\mathbf{s}_t .
\label{eq:physics_delta}
\end{equation}
The residual input is formed as
\begin{equation}
\mathbf{q}_t
=
\left[
\mathbf{c}_t,
\tilde{\boldsymbol{\theta}}_t,
\mathbf{h}_t,
\hat{\mathbf{s}}^{\mathrm{phy}}_{t+1},
\Delta\hat{\mathbf{s}}^{\mathrm{phy}}_t
\right],
\label{eq:residual_input}
\end{equation}
and the residual network produces
\begin{equation}
\mathbf{r}^{\mathrm{corr}}_t
=
r_{\psi}(\mathbf{q}_t).
\label{eq:residual_output}
\end{equation}
Here, $\mathbf{r}^{\mathrm{corr}}_t$ denotes the normalized residual-network
output. It is not directly the physical correction added to the state. The
actual gated and scaled correction is
\begin{equation}
\Delta\hat{\mathbf{s}}^{\mathrm{corr}}_t
=
\hat{\mathbf{s}}_{t+1}
-
\hat{\mathbf{s}}^{\mathrm{phy}}_{t+1}
=
\lambda_r
\mathbf{s}_{\Delta}
\odot
\mathbf{g}_r
\odot
\mathbf{r}^{\mathrm{corr}}_t ,
\label{eq:gated_residual_correction}
\end{equation}
where $\mathbf{s}_{\Delta}$ is the state-wise correction scale,
$\lambda_r$ controls the residual magnitude, and $\mathbf{g}_r$ is a sigmoid
residual gate. The final prediction is then
\begin{equation}
\hat{\mathbf{s}}_{t+1}
=
\hat{\mathbf{s}}^{\mathrm{phy}}_{t+1}
+
\Delta\hat{\mathbf{s}}^{\mathrm{corr}}_t .
\label{eq:final_prediction}
\end{equation}

Here, $\mathbf{c}_t$ is the physics-conditioned latent feature,
$\tilde{\boldsymbol{\theta}}_t$ is the normalized parameter deviation, and
$\mathbf{h}_t$ is the latest raw state-action input. The residual input
$\mathbf{q}_t$ therefore combines latent context, normalized parameter
adaptation, the latest state-action input, the physics prediction, and the
physics-predicted state increment. This design makes the residual branch
secondary to the analytical dynamics layer: the network first predicts the next
dynamic state through $f_{\mathrm{dyn}}(\cdot)$ using bounded physical
parameters, and the residual pathway only refines the remaining local prediction
error.

\subsection{State Propagation}
\label{subsec:state_propagation}

The physics-guided dynamics layer uses the most recent actuator feedback and
command signals to construct the applied steering and throttle inputs:
\begin{equation}
    \delta_t
    =
    \delta_{\mathrm{fb},t}
    +
    \delta_{\mathrm{cmd},t},
    \qquad
    T_t
    =
    T_{\mathrm{fb},t}
    +
    T_{\mathrm{cmd},t}.
\label{eq:applied_control}
\end{equation}
Although the analytical dynamics step uses the current applied control, temporal
actuator effects are captured implicitly through the SSM-GRU history encoder,
which receives both feedback and command histories as part of the input window.
This allows the learned adaptation vector to reflect actuator lag, transient
response, and recent control variation without introducing a separate
hand-designed actuator-delay model.

After computing the longitudinal acceleration $a_x$, lateral acceleration
$a_y$, and yaw acceleration $\dot{\omega}$, the physics-based next-state
prediction is obtained using a forward Euler step:
\begin{equation}
    \hat{\mathbf{x}}^{\mathrm{phy}}_{t+1}
    =
    \mathbf{x}_t
    +
    T_s
    \begin{bmatrix}
    a_x \\
    a_y \\
    \dot{\omega}
    \end{bmatrix}.
\label{eq:state_propagation}
\end{equation}
The sampling time $T_s$ is selected according to the experimental platform. The
ORCA-scale model uses the small-scale vehicle sampling time and parameter
bounds, while the IAC-scale model uses the corresponding full-scale vehicle
configuration. This keeps the propagation structure unchanged while allowing
the learned parameter adaptation to account for platform-specific dynamics.
\section{Data Generation and Learned Coefficients} \label{sec:datagenerration}

\subsection{Training and Testing Data}

To evaluate the predictive performance and computational efficiency of the proposed \method{}, experiments were conducted on two datasets representing different autonomous racing regimes: an ORCA-scale simulated dataset, denoted by $\SD{}$, and a full-scale IAC dataset, denoted by $\RD{}$.
For fair comparison, all models were trained on the same training subset and evaluated on the same unseen test subset within each dataset.

The simulated dataset $\SD{}$ was generated using a 1:43-scale autonomous racecar simulation environment~\cite{jain2020bayesrace}.
The data were sampled at 50,Hz while a pure-pursuit controller tracked a predefined racing line.
Two tracks were used to evaluate cross-track generalization: Track~1 (ETHZ) was used to generate the training subset $\SDtr{}$, while Track~2 (ETHZMobil) was reserved exclusively as the unseen test subset $\SDte{}$.

The real-world dataset $\RD{}$ was collected from a full-scale autonomous Indy racecar platform~\cite{kulkarni2023racecar}.
Vehicle states were obtained from an Extended Kalman Filter (EKF)-based state-estimation pipeline, which fused measurements from dual RTK-GNSS receivers and onboard inertial measurement units (IMUs).
The training subset $\RDtr{}$ consists of 62,419 samples recorded over three laps at the Putnam Park Road Course in Indiana, while the unseen test subset $\RDte{}$ contains 56,602 samples collected over two laps at the Las Vegas Motor Speedway.
This split evaluates whether the learned dynamics model can generalize across different full-scale racing environments rather than only interpolate within a single track.

Model performance was first evaluated using one-step prediction metrics.
Specifically, we computed the root mean square error (RMSE) and maximum absolute error ($\epsilon_{\max}$) for the dynamic vehicle states $[\vx, \vy, \w]^\top$.
Beyond single-step prediction, we also assessed multi-step forecasting performance by recursively rolling out each learned dynamics model over a fixed prediction horizon.
From these rollouts, we computed the average displacement error (ADE) and final displacement error (FDE), consistent with trajectory forecasting benchmarks such as~\cite{alahi2016social}.
A 300 ms prediction horizon was used for the simulated dataset, while a 600 ms horizon was used for the real-world dataset.

Finally, model complexity was compared using the number of trainable parameters, floating-point operations (FLOPs, reported in millions per sample), and average inference time.
All experiments were performed on a workstation equipped with an Intel Core i9-14900K processor, an NVIDIA GeForce RTX 4070 SUPER GPU, and 64 GB of DDR5 memory.

\subsection{Learned Parameter Bounds}
\label{subsec:learned_parameter_bounds}

\begin{table}[t]
\caption{Coefficient Ranges and Learned Parameters for the Single-Track Model on Simulated and Real Track Data}
\label{tab:coeff_stats_combined}
\centering
\scriptsize
\renewcommand{\arraystretch}{1.15}
\resizebox{\columnwidth}{!}{%
\begin{tblr}{
width = \textwidth,
colspec = {c c c c c},
row{1-2} = {bg=blue!12, fg=black, font=\bfseries},
colsep = 1.2mm,
rowsep = 0.6mm,
vlines,
hlines,
}

\SetCell[r=2]{c} Coefficient
& \SetCell[c=2]{c} Simulated Track Data & 
& \SetCell[c=2]{c} Real Track Data
& \\

& \textbf{Min--Max Range} & \textbf{\method{}}
& \textbf{Min--Max Range} & \textbf{\method{}} \\

$B_f$ & $5.0$--$30.0$ & $8.35 \pm 0.20$
& $5.0$--$30.0$ & $6.41 \pm 3.17$ \\

$C_f$ & $0.5$--$2.0$ & $0.508 \pm 4.6{\times}10^{-4}$
& $0.5$--$2.0$ & $0.73 \pm 0.26$ \\

$D_f$ & $0.1$--$0.9$ & $0.175 \pm 0.012$
& $100$--$10000$ & $5416 \pm 2032$ \\

$E_f$ & $-2.0$--$0.0$ & $-0.471 \pm 0.008$
& $-2.0$--$0.0$ & $-0.04 \pm 0.04$ \\

$B_r$ & $5.0$--$30.0$ & $7.74 \pm 0.19$
& $5.0$--$30.0$ & $7.08 \pm 3.89$ \\

$C_r$ & $0.5$--$2.0$ & $0.569 \pm 0.011$
& $0.5$--$2.0$ & $0.62 \pm 0.20$ \\

$D_r$ & $0.1$--$0.9$ & $0.119 \pm 0.001$
& $100$--$10000$ & $6001 \pm 2498$ \\

$E_r$ & $-2.0$--$0.0$ & $-0.422 \pm 0.099$
& $-2.0$--$0.0$ & $-1.77 \pm 0.28$ \\

$I_z$ & $1.4{\times}10^{-5}$--$5.6{\times}10^{-5}$
& $2.79{\times}10^{-5} \pm 5.3{\times}10^{-7}$
& $500$--$2000$ & $1999.7 \pm 0.8$ \\

$S_{hf}$ & $-0.02$--$0.02$ & $-0.0032 \pm 7.7{\times}10^{-4}$
& $-0.02$--$0.02$ & $0.0028 \pm 0.0089$ \\

$S_{vf}$ & $-0.003$--$0.003$ & $0.0016 \pm 5.0{\times}10^{-4}$
& $-300$--$300$ & $159.2 \pm 179.6$ \\

$S_{hr}$ & $-0.02$--$0.02$ & $-0.0025 \pm 4.7{\times}10^{-4}$
& $-0.02$--$0.02$ & $0.0049 \pm 0.0089$ \\

$S_{vr}$ & $-0.003$--$0.003$ & $-3.9{\times}10^{-4} \pm 3.3{\times}10^{-4}$
& $-300$--$300$ & $163.3 \pm 180.6$ \\

\end{tblr}
}
\end{table}

Table~\ref{tab:coeff_stats_combined} summarizes the admissible coefficient
ranges and learned parameter statistics used by the physics-guided dynamics
layer of \method{}. 
These parameters correspond to the adaptive dynamics vector
$\boldsymbol{\theta}_t$ introduced in Eq.~\eqref{eq:bounded_param_adapt}. 
The adapted parameter set includes Pacejka-style tire coefficients
$(B_f,C_f,D_f,E_f)$ and $(B_r,C_r,D_r,E_r)$, front and rear slip-angle shifts
$(S_{hf},S_{hr})$, lateral force shifts $(S_{vf},S_{vr})$, longitudinal
drive-force coefficients $(C_{m1},C_{m2})$, rolling resistance coefficients
$(C_{r0},C_{r2})$, and yaw moment of inertia $I_z$. 
Together, these quantities govern lateral tire-force generation, 
load-dependent force scaling, and yaw dynamics inside the analytical vehicle
model. 
The parameter bounds are selected to define physically admissible
operating ranges for the corresponding platform, while the learned statistics
show how the model uses these ranges during prediction.

\begin{figure}[h]
    \centering
    \includegraphics[width=1\linewidth]{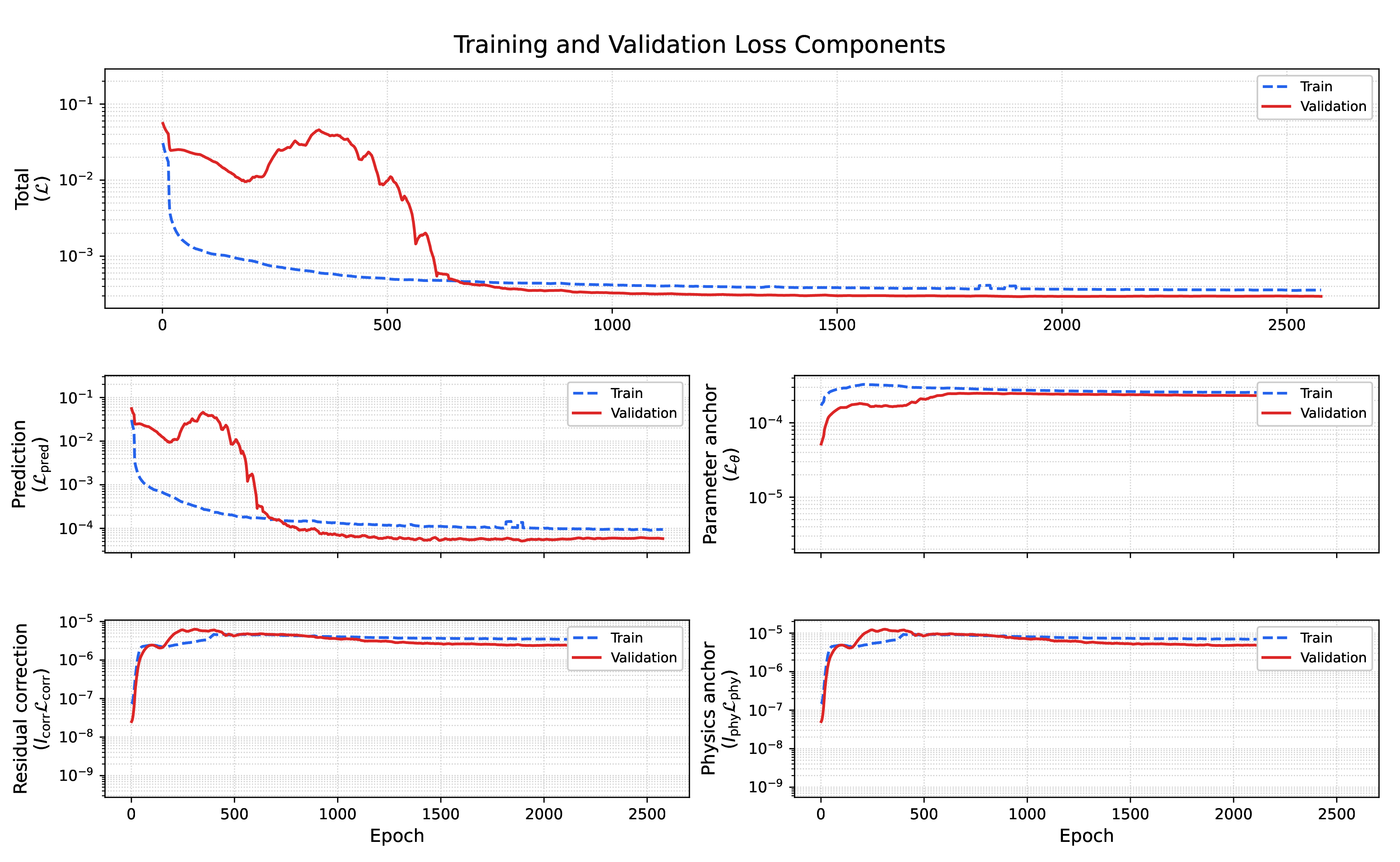}
    \caption{Evolution of training and validation loss components defined in Eq.(\ref{eq:total_loss}). 
    }
    \label{fig:loss_components}
\end{figure}
\begin{table*}[b]
\caption{Ablation Study of the Proposed \method{} Architecture on the real-world evaluation subset}
\label{tab:ablation}
\centering
\begin{tblr}{
  width = \linewidth,
  colspec = {
    X[1.0,c] X[1.3,c] X[1.5,c]
    X[0.45,c] X[0.45,c] X[0.55,c] X[0.55,c]
    X[0.80,c] X[0.70,c] X[0.60,c]
    X[0.80,c] X[0.80,c] X[0.80,c]
  },
  row{2} = {font=\bfseries\scriptsize},
  hlines = {1pt, black},
  hline{4,5,6,7,8} = {dashed,fg=gray},
  vlines = {0.6pt, black},
  rowsep = 1pt,
  colsep = 1pt,
  stretch = 0.9,
   row{9}={bg=blue!12},
}
\SetCell[r=2]{c}\textbf{Variant}
& \SetCell[r=2]{c}\textbf{Ablated Component}
& \SetCell[r=2]{c}\textbf{Role Tested}
& \SetCell[c=4]{c}\textbf{Components}
& & & 
& \SetCell[c=3]{c}\textbf{Efficiency}
& & 
& \SetCell[c=3]{c}\textbf{Single Lap RMSE}
& & \\
& & 
& \textbf{SSM Enc.} & \textbf{GRU Enc.} & \textbf{Param. Head} & \textbf{Residual Head}
& \textbf{Active Params. (M)} & \textbf{FLOPs (M)} & \textbf{Time(s)}
& { \textbf{$v_x$}\\\textbf{($\mathrm{m\,s^{-1}}$)} }
& { \textbf{$v_y$}\\\textbf{($\mathrm{m\,s^{-1}}$)} }
& { \textbf{$\omega$}\\\textbf{($\mathrm{rad\,s^{-1}}$)} } \\

Physics-Only STM
& No learned adaptation
& Data adaptation
& \xmark & \xmark & \xmark & \xmark
& $0.0000$ & $4.0{\times}10^{-5}$ & $0.001$
& $3.50{\times}10^{-2}$ & $1.24{\times}10^{-1}$ & $4.37{\times}10^{-2}$ \\

Neural-only predictor
& No physics layer
& Physics guidance
& \cmark & \cmark & \xmark & \xmark
& $0.0756$ & $0.1770$ & $0.002$
& $2.67{\times}10^{-2}$ & $1.23{\times}10^{-1}$ & $6.55{\times}10^{-2}$ \\

\method{} w/o SSM 
& Remove SSM memory
& Long-horizon memory
& \xmark & \cmark & \cmark & \cmark
& $0.0795$ & $0.1398$ & $0.002$
& $2.86{\times}10^{-2}$ & $2.58{\times}10^{-2}$ & $3.60{\times}10^{-2}$ \\

\method{} w/o GRU
& Remove GRU adapter
& Transient response
& \cmark & \xmark & \cmark & \cmark
& $0.0729$ & $0.1838$ & $0.002$
& $2.72{\times}10^{-2}$ & $3.78{\times}10^{-2}$ & $7.25{\times}10^{-3}$ \\

\method{} w/o Param. Adapt.
& Fixed physics parameters
& Parameter adaptation
& \cmark & \cmark & \xmark & \cmark
& $0.0790$ & $0.1794$ & $0.002$
& $2.93{\times}10^{-2}$ & $8.20{\times}10^{-2}$ & $1.01{\times}10^{-2}$ \\

\method{} w/o Residual
& Remove residual head
& Unmodeled dynamics
& \cmark & \cmark & \cmark & \xmark
& $0.0516$ & $0.1257$ & $0.002$
& $2.58{\times}10^{-2}$ & $2.46{\times}10^{-2}$ & $2.31{\times}10^{-2}$ \\

\method{}
& Full model
& Complete framework
& \cmark & \cmark & \cmark & \cmark
& $0.0811$ & $0.1838$ & $0.002$
& $2.46{\times}10^{-2}$ & $1.90{\times}10^{-2}$ & $7.22{\times}10^{-3}$ \\
\end{tblr}
\end{table*}
Rather than allowing the network to predict unconstrained dynamics coefficients, \method{} maps the learned temporal representation to physically meaningful vehicle parameters and clips the resulting values to predefined lower and upper bounds.
This bounded parameterization acts as a physics guardrail during learning and
inference. 
It prevents the network conditioning from producing unrealistic
coefficient estimates while still allowing the internal dynamics model to adapt to local operating conditions. 
In particular, the temporal encoder captures
recent state, actuator-feedback, command history and the parameter head then
uses this context to generate the adapted vector $\boldsymbol{\theta}_t$ for the current prediction step. 
As a result, the model does not rely on a single fixed
set of dynamics coefficients, but also does not discard the structure of the
single-track dynamics model.

The variations in the tire and slip/force-shift parameters primarily influence lateral force generation, whereas changes in the drive and resistance coefficients affect longitudinal vehicle response, and variations in $I_z$ influences rotational dynamics while remaining constrained within physically admissible limits. 
These trends 
are further validated by Fig.~\ref{fig:loss_components} that shows the evolution of the training and validation loss components from Eq.~(\ref{eq:total_loss}) during \method{} optimization. 
The learned parameter statistics reported in Table.~\ref{tab:coeff_stats_combined} along with Fig.~\ref{fig:loss_components} demonstrate that the model adapts through physically meaningful channels rather than relying solely on latent feature corrections.

%

\section{Results} \label{sec:Results}
This section evaluates \method{} performance in both open-loop prediction and closed-loop control settings through component-wise ablation and comparisons with representative state-of-the-art baselines, providing insight into the contribution of each architectural component and the overall performance of the proposed \method{} framework.

\subsection{Open-loop}
\subsubsection{Ablation Study}
\label{subsec:ablation_study}
To quantify the contribution of each module in \method{}, we conduct a component-wise ablation study on the real-world evaluation subset. 
Each variant is trained using the real-world train set $\RDtr{}$ and evaluated on the held-out real-world test set $\RDte{}$. 
The results are summarized in Table~\ref{tab:ablation}. 
These ablations are designed to isolate the roles of SSM encoder, GRU encoder, bounded parameter head, and gated residual head.
%
In addition to single-lap prediction error, we report active parameter count, FLOPs, and inference time to evaluate the accuracy-efficiency trade-off of each variant.

The Physics-Only STM denotes the nominal single-track model evaluated without any of the above 4 mentioned components. 
This variant has essentially no learned parameters and the lowest computational cost, but it exhibits substantially larger prediction errors, particularly in $v_y$ and $\omega$. 
This confirms that the fixed analytical dynamics model alone is insufficient to capture the lateral and rotational behavior present in
real-world racing data. 
Conversely, the neural-only predictor removes the structured dynamics pathway and relies on learned temporal prediction alone.
Although this model is lightweight and can fit some short-horizon trends, it produces significantly larger errors than the full \method{} model and does not provide physics-guided state propagation, bounded parameter adaptation, or interpretable intermediate dynamics.

The temporal encoder ablations show that the SSM and GRU branches provide complementary information. \method{} w/o SSM variant shows an increased yaw-rate RMSE relative to the full model, indicating that the state-space branch contributes useful longer-memory temporal structure for rotational dynamics. 
\method{} w/o GRU encoder leads to a larger degradation in lateral velocity RMSE, showing that the recurrent branch is important for capturing short-term maneuver-dependent transients in the recent state-action history. These results support the use of
parallel SSM and GRU encoders rather than relying on only one temporal pathway.

The \method{} w/o parameter adaptation demonstrates the importance of bounded physical adaptation. When the parameter adaptation head is removed, the model can no longer adjust the dynamics parameters according to the encoded operating
condition, which results in a large increase in $v_y$ RMSE. 
This suggests that adaptive parameter conditioning is especially important for representing changes in lateral operating regimes, tire-force response, and load-dependent behavior. 
Similarly, \method{} w/o residual head degrades both lateral velocity and yaw-rate prediction compared with the full model. 
This indicates that, even with bounded adaptive parameters, a small learned correction remains useful for capturing local effects not fully represented by the analytical dynamics layer, such as actuator-response mismatch, sensing noise, and unmodeled surface
variation.

\begin{table*}[t]
\caption{Open-Loop Model Performance Comparison with SoTA models}
\label{tab:open_loop}
\centering
\begin{tblr}{
  width = \linewidth,
  colspec = {X[0.4,l] X[1.0,c] *{10}{X[0.83,c]} X[0.7,c]} ,
  row{2} = {font=\bfseries\scriptsize},
  hlines = {1pt, black},
  hline{4,5, 7,8} = {dashed,fg=gray},
  vlines = {0.6pt, black},
  rowsep = 1pt,
  colsep = 1pt,
  stretch = 0.9,
  row{5,8}={bg=blue!12},
}
\SetCell[r=2]{c} &  \SetCell[r=2]{c} \textbf{Method}
& \SetCell[c=2]{c} $\boldsymbol{v_x}$ (\si{\meter\per\second}) &
& \SetCell[c=2]{c} $\boldsymbol{v_y}$ (\si{\meter\per\second}) &
& \SetCell[c=2]{c} $\boldsymbol{\omega}$ (\si{\radian\per\second}) &
& \SetCell[c=2]{c} \textbf{Displacement} (\si{\meter}) &
& \SetCell[r=2]{c} {\textbf{Total}\\\textbf{Parameters}} 
& \SetCell[r=2]{c} {\textbf{FLOPs} \\ \textbf{(M)}} 
& \SetCell[r=2]{c} \textbf{Avg.}\ \textbf{Inference Time} (\si{\second}) \\
& 
& RMSE & $\boldsymbol{\epsilon_{\max}}$
& RMSE & $\boldsymbol{\epsilon_{\max}}$
& RMSE & $\boldsymbol{\epsilon_{\max}}$
& ADE & FDE & & \\

\SetCell[r=3]{c} $\SDte{}$
& DDM
& $1.73{\times}10^{-4}$ & $1.34{\times}10^{-3}$
& $2.93{\times}10^{-3}$ & $1.49{\times}10^{-2}$
& $1.09{\times}10^{-1}$ & $6.41{\times}10^{-1}$
& \num{7.59e-03} & \num{1.60e-03}
& $2.88 $ M
& $5.75 $
& $0.0018$ \\

& FTHD
& $1.70{\times}10^{-4}$ & $1.18{\times}10^{-3}$
& $1.68{\times}10^{-3}$ & $3.35{\times}10^{-2}$
& $4.83{\times}10^{-1}$ & $2.89{\times}10^{-1}$
& \num{6.64e-02} & \num{1.78e-01}
& $2.88 $ M
& $5.79 $
& $0.0019$ \\

& \method{}
& $3.79{\times}10^{-4}$ & $2.97{\times}10^{-3}$
& $2.09{\times}10^{-3}$ & $2.51{\times}10^{-2}$
& $1.14{\times}10^{-1}$ & $1.93{\times}10^{-1}$
& \num{2.25e-03} & \num{3.75e-03}
& $0.42$ M
& $1.11$ 
& $0.0002$ \\

\SetCell[r=3]{c} $\RDte{}$
& \ddm{}
& $3.38{\times}10^{-2}$ & $9.46{\times}10^{-1}$
& $7.29{\times}10^{-2}$ & $4.95{\times}10^{-1}$
& $1.56{\times}10^{-2}$ & $2.15{\times}10^{-1}$
& \num{1.55e-01} & \num{3.02e-01}
& $0.085$ M
& $0.19$ 
& $0.003$ \\

& FTHD
& $2.70{\times}10^{-2}$ & $9.38{\times}10^{-1}$
& $3.38{\times}10^{-2}$ & $4.15{\times}10^{-1}$
& $6.45{\times}10^{-2}$ & $1.29{\times}10^{-1}$
& \num{1.43e-01} & \num{3.32e-01}
& $0.085$ M
& $0.19$ 
& $0.003$ \\

& \method{}
& $2.46{\times}10^{-2}$ & $9.27{\times}10^{-1}$
& $1.90{\times}10^{-2}$ & $9.47{\times}10^{-2}$
& $7.22{\times}10^{-3}$ & $2.85{\times}10^{-2}$
& \num{1.51e-01} & \num{2.46e-01}
& $0.081$ M
& $0.18$ 
& $0.002$ 
\end{tblr}
\end{table*}

The complete \method{} model combines the SSM encoder, GRU encoder, bounded parameter head, and gated residual head. 
It achieves the lowest RMSE in $v_x$, $v_y$, and $\omega$ among the evaluated variants while maintaining real-time inference with only a modest computational cost. 
%
%
Overall, the ablation study shows that each component contributes a distinct function, and that the full \method{} architecture provides the strongest balance of predictive accuracy, temporal adaptability, physical interpretability, and computational efficiency on real-world data.

\subsubsection{\texorpdfstring{\sota{} Models}{SoTA Models}} \label{subsubsec:sota_models}

Table~\ref{tab:open_loop} compares the open-loop predictive performance of the proposed \method{} against representative state-of-the-art vehicle dynamics models (\ddm{} \cite{chrosniak2024deep}, \fthd{} \cite{fang2025fine}) on both simulated and real-world datasets. 
The simulated evaluation uses held-out simulated track data to assess generalization under controlled conditions, while the real-world evaluation uses a held-out subset collected from the full-scale autonomous Indy racecar. 
Each model is trained and tested using the corresponding subsets described in Sec.~\ref{sec:datagenerration}.
Performance is evaluated using one-step prediction errors for $v_x$, $v_y$, and $\omega$, displacement-level roll out metrics, model size, computational complexity, and average inference time.

On the simulated test set $\SDte{}$, \method{} offers a favorable balance between prediction accuracy and computational efficiency. 
%
%
Prediction performance remains competitive, with a $28.7\%$ reduction in lateral velocity RMSE relative to \ddm{} and a $76.4\%$ reduction in yaw-rate RMSE relative to \fthd{}. 
Although \method{} does not achieve the lowest error across every prediction channel, particularly in longitudinal velocity, the overall results indicate that competitive predictive performance can be maintained with a substantially more compact model.

On the real-world test set $\RDte{}$, the advantage of \method{} becomes more pronounced. 
Compared with \ddm{} and \fthd{}, \method{} reduces RMSE across $[v_x,v_y,\omega]^T$ while also improving displacement-level rollout metrics.
These improvements suggest that the combination of temporal memory, physics-guided adaptation, and residual refinement enables better generalization to previously unseen operating conditions encountered by the full-scale racecar.
%
%
%

Beyond prediction accuracy, Table.~\ref{tab:open_loop} highlights the computational efficiency of \method{}. 
Across both simulated and real-world datasets, \method{} consistently achieves lower parameter counts, reduced computational complexity, and the lowest inference latency relative to \ddm{} and \fthd{}. 
While the proposed approach is not uniformly optimal across all open-loop metrics, it delivers the strongest overall trade-off between predictive accuracy, model compactness, and computational efficiency. 
These properties are particularly desirable for autonomous racing applications, where the learned dynamics model must support repeated real-time evaluations within model predictive controllers.


\subsection{Closed-loop} \label{sec:closed_loop_results}
To further assess the effect of each architectural component beyond one-step prediction accuracy and evaluate whether the open-loop prediction performance translates to closed-loop, we perform: 1) A closed-loop ablation study (\ref{sec:cl_ablation}) and 2) A comparison \method{} with representative \sota{} dynamic models \ddm{} and \fthd{} (\ref{sec:cl_sota}) using a nonlinear model predictive controller (NMPC).  
\begin{figure}[h]
\centering
\includegraphics[width=\linewidth]{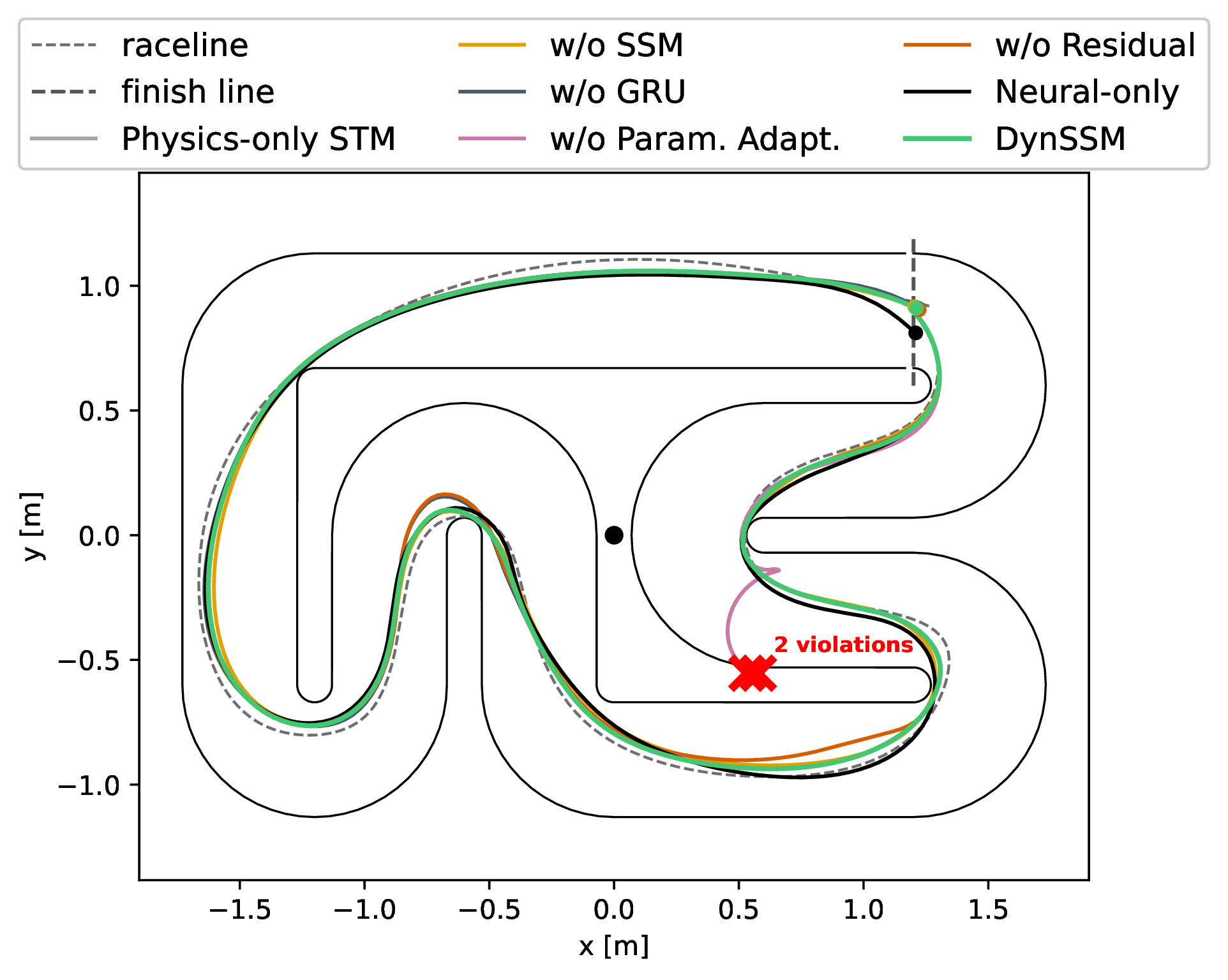}
\caption{Closed-loop trajectories of \method{} ablation variants on the
ETHZMobil track. Variants that violate the track illustrate the importance
of learned adaptation for feasible closed-loop operation.}
\label{fig:closed_loop_ablation}
\end{figure}

\begin{figure*}[hb]
\centering
\includegraphics[width=\linewidth]{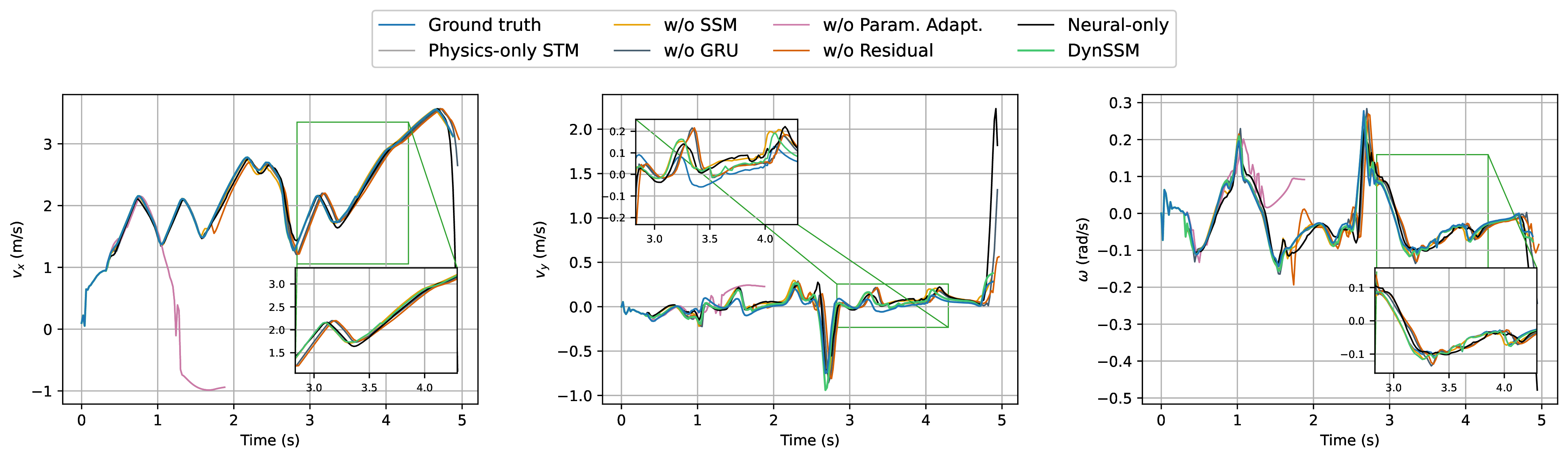}
\caption{Closed-loop state-profile comparison of the \method{} ablation variants on the ETHZMobil track. 
}
\label{fig:closed_loop_ablation_diagnostics}
\end{figure*}

\subsubsection{Ablation Study} \label{sec:cl_ablation}
\begin{figure}[h]
\centering
\includegraphics[width=\columnwidth]{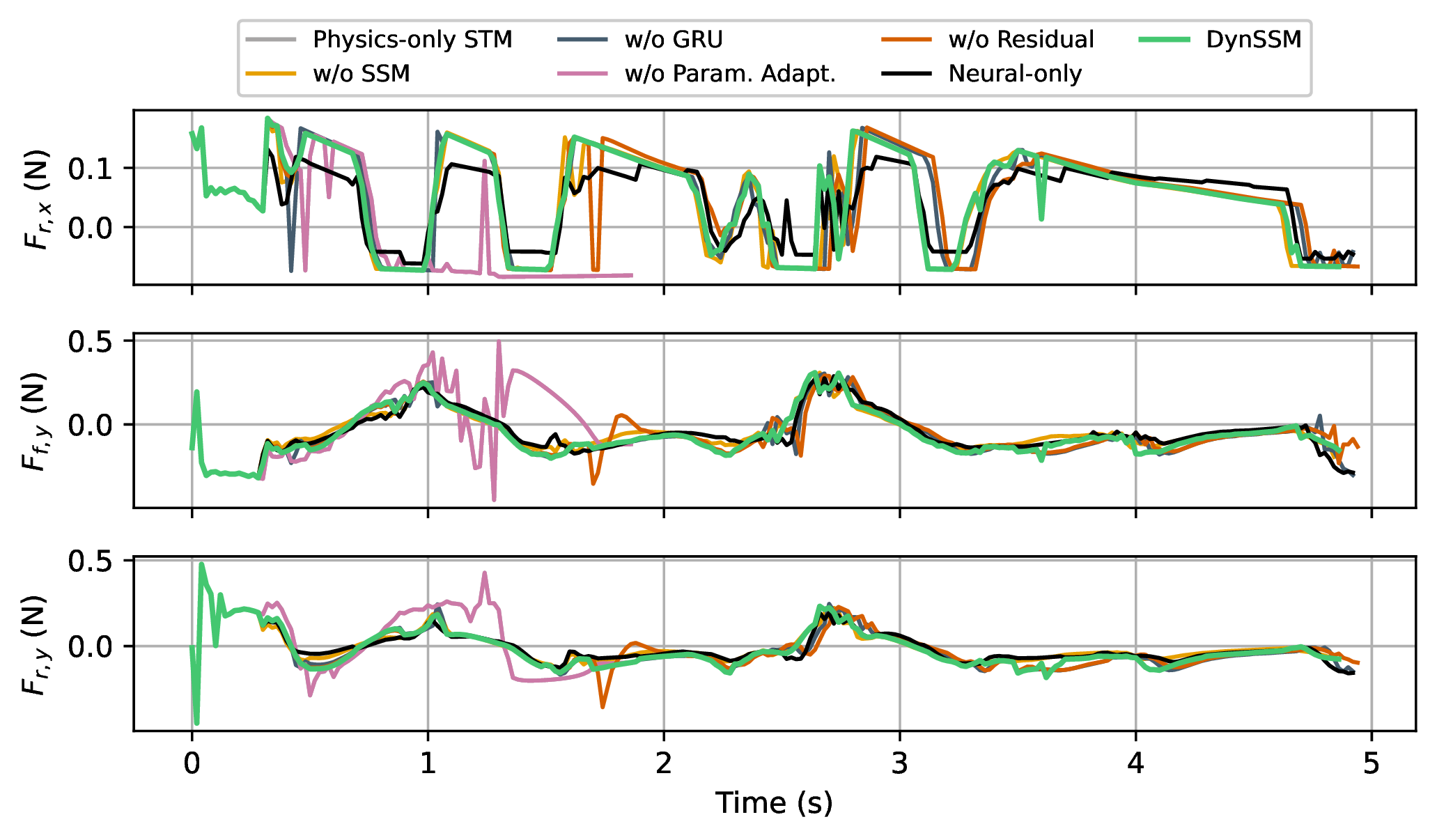}
\caption{Estimated force profiles for the \method{} ablation variants on
the ETHZMobil track.}
\label{fig:cl_forces}
\end{figure}

The closed-loop ablation results are summarized in Table~\ref{tab:closed_loop_ablation} and Figs.~\ref{fig:closed_loop_ablation}-\ref{fig:cl_forces}. 
Unlike the open-loop ablation, this evaluation examines how each architectural component influences controller feasibility and performance when the learned dynamics model is embedded within the NMPC architecture. 
Performance is assessed using lap time, average velocity, and steering total variation ($\mathrm{TV}_{\delta}$), where lower values of $\mathrm{TV}_{\delta}$ indicate smoother steering behavior - lower values
indicate smoother steering behavior.

Fig.~\ref{fig:closed_loop_ablation} shows the closed-loop trajectory during one lap, Fig.~\ref{fig:closed_loop_ablation_diagnostics} and Fig.~\ref{fig:cl_forces} show the state evolution of $\mathbf{s}_t$ and estimated force profiles respectively over one lap for all the ablation variants. 
The results show that bounded physics-guided adaptation is essential for feasible closed-loop operation. 
Both the Physics-Only STM and the \method{} w/o parameter adaptation exhibit degraded state evolution that cannot be adequately corrected by the controller, ultimately leading to track violations. 
This demonstrates that a fixed analytical model is insufficient for aggressive racing maneuvers and that adaptive parameter conditioning plays a critical role in maintaining predictive fidelity under changing operating conditions. 
In contrast, all variants containing the adaptation mechanism successfully complete the lap.
\begin{table}[t]
\caption{Closed-loop ablation study of the proposed \method{} architecture on the ETHZMobil track.}
\label{tab:closed_loop_ablation}
\centering
\resizebox{\columnwidth}{!}{%
\begin{tblr}{
  width = \linewidth,
  colspec = {
    X[2.0,c]
    X[0.75,c]
    X[0.9,c]
    X[0.9,c]
    X[0.9,c]
  },
  row{1} = {font=\bfseries\scriptsize},
  rows = {font=\scriptsize},
  hlines = {1pt, black},
  hline{3,4,5,6,7} = {dashed,fg=gray},
  vlines = {0.6pt, black},
  rowsep = 1pt,
  colsep = 1pt,
  stretch = 0.85,
   row{8}={bg=blue!12},
}
\textbf{Variant}
& \textbf{Violation}
& \textbf{Lap Time} (\si{\second})
& $\boldsymbol{\bar{v}_x}$ (\si{\meter\per\second})
& \textbf{Steering TV} \\

Physics-Only STM
& \cmark
& $1.88$
& $0.57$
& $2.11$ \\

Neural-only predictor
& \xmark
& $4.94$
& $2.11$
& $4.79$ \\

\method{} w/o SSM
& \xmark
& $4.86$
& $2.14$
& $4.01$ \\

\method{} w/o GRU
& \xmark
& $4.94$
& $2.15$
& $4.52$ \\

\method{} w/o Param. Adapt.
& \cmark
& $1.88$
& $0.57$
& $2.11$ \\

\method{} w/o Residual
& \xmark
& $4.96$
& $2.14$
& $5.00$ \\

\method{}
& \xmark
& $4.88$
& $2.15$
& $4.08$ \\
\end{tblr}}
\end{table}

Among the feasible learned models, the \method{} w/o SSM variant achieves the shortest lap time, suggesting that the GRU pathway captures sufficient short-term temporal information for this particular track. 
Nevertheless, the full \method{} architecture achieves comparable performance while retaining the SSM branch, which is designed to model longer-horizon temporal dependencies and operating-condition history. 
The residual refinement mechanism primarily affects control smoothness, removing it preserves feasibility but increases steering total variation, indicating less consistent controller behavior.

%

\subsubsection{\texorpdfstring{\sota{} Models}{SoTA Models}}\label{sec:cl_sota}
In the \sota{} closed-loop comparison, the NMPC configuration is kept identical across all dynamics models, including the prediction horizon, cost weights, constraints, solver settings, and reference raceline. 
Consequently, any differences in lap completion, vehicle behavior, and controller performance can be attributed to the predictive dynamics model embedded within the NMPC optimization.

\begin{figure*}[hb]
    \centering
    \includegraphics[width=1\linewidth]{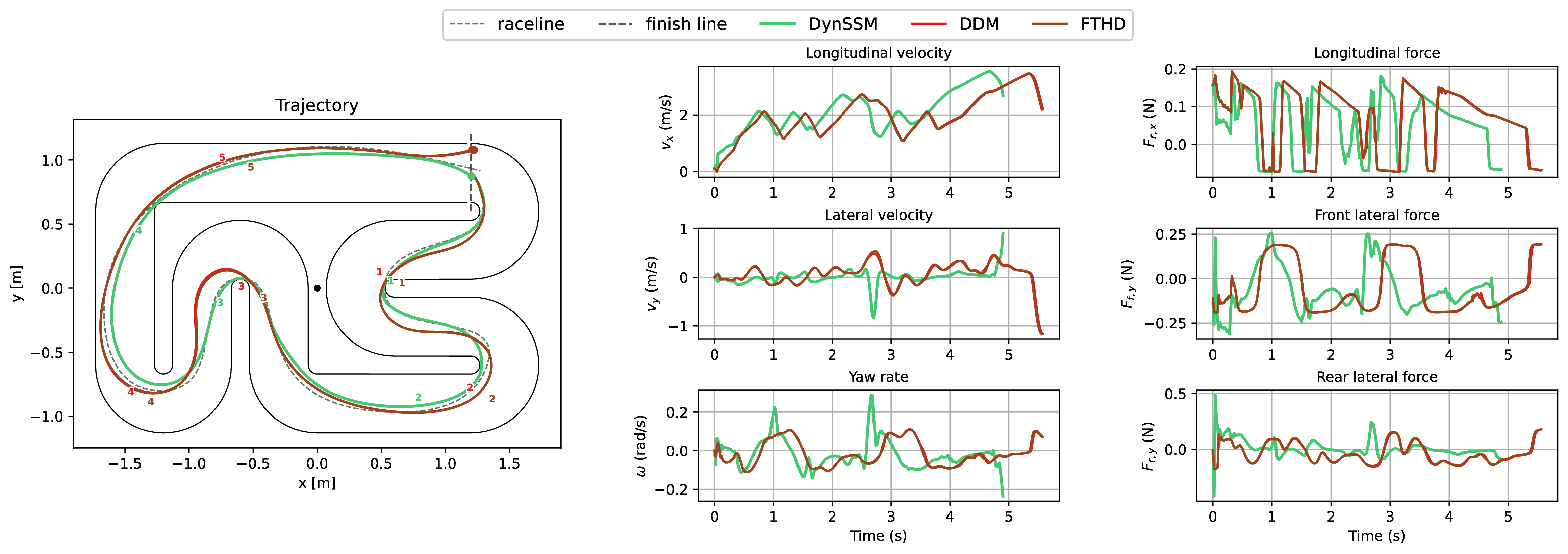}
    \caption{Closed-loop diagnostic comparison of \method{} with \sota{} dynamics models on the ETHZMobil track. The trajectory plot (left most) shows the closed-loop path followed during NMPC rollout, while the state (middle) show the model-dependent vehicle-state evolution and and force profiles (right) show the estimated tire-force behavior used by the controller.}
    \label{fig:closed_loop_sota_diagnostics}
\end{figure*}

\begin{table}[H]
\caption{Closed-loop simulation results}
\label{tab:closed_loop_sota}
\begin{tblr}{
  width=\columnwidth,
  colspec={X[1.0,c,m] X[1.0,c,m] X[0.85,c,m] X[0.82,c,m] X[0.91,c,m] X[0.97,c,m] X[1.1,c,m]},
  row{1}={font=\bfseries},
  hline{1,2,5,8,11}={1pt,black},
  hline{3,4,6,7,9,10}={2-Z}{dashed,fg=gray},
  vlines={0.6pt,black},
  cell{2}{1}={r=3}{c},
  cell{5}{1}={r=3}{c},
  cell{8}{1}={r=3}{c},
  row{4,7,10}={bg=blue!12},
}
Track & Model & Lap Time (\si{\second}) & $v_x$ (\si{\meter\per\second}) &
$v_y$ (\si{\meter\per\second}) & $\omega$ (\si{\radian\per\second}) & \textbf{Violation} \\

{ETHZ\\Mobil} 
    & DDM     & 5.58   & 1.98 & 0.18 & 3.37 & \xmark \\
    & FTHD    & 5.62     & 1.96 & 0.16 & 3.27 & \xmark \\
    & \method{}  & 4.88   & 2.15 & 0.08 & 3.47 & \xmark \\

Austin    
    & DDM     & 106.40 & 3.82 & 0.05 & 0.31 & \xmark \\
    & FTHD    & 106.44 & 3.81 & 0.05 & 0.30 & \xmark \\
    & \method{}  & 104.06 & 3.86 & 0.04 & 0.30 & \xmark \\

IMS       
    & DDM     & 69.72  & 4.16 & 0.04 & 0.09 & \xmark \\
    & FTHD    & 69.72 & 4.15 & 0.05 & 0.09 & \xmark \\
    & \method{} & 67.46  & 4.78 & 0.07 & 0.12 & \xmark \\

\end{tblr}
\end{table}

The quantitative results are summarized in Table~\ref{tab:closed_loop_sota}, while representative diagnostic rollouts on the ETHZMobil track are shown in Fig.~\ref{fig:closed_loop_sota_diagnostics}. 
On ETHZMobil, all evaluated \sota{} models successfully complete the lap. 
%
\begin{figure}[ht]
    \centering
    \includegraphics[width=1\columnwidth]{ 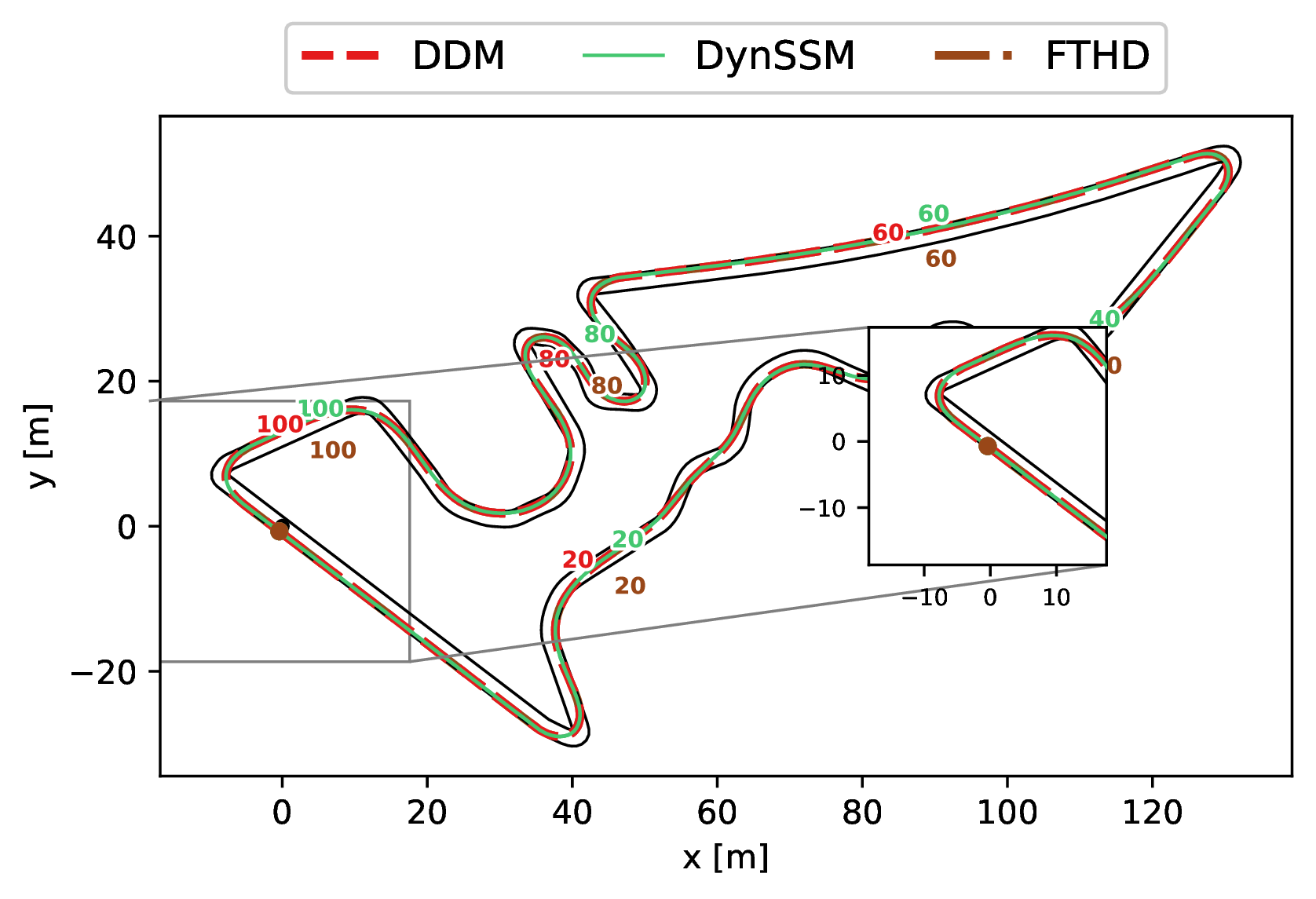}
    \includegraphics[width=1\columnwidth]{ 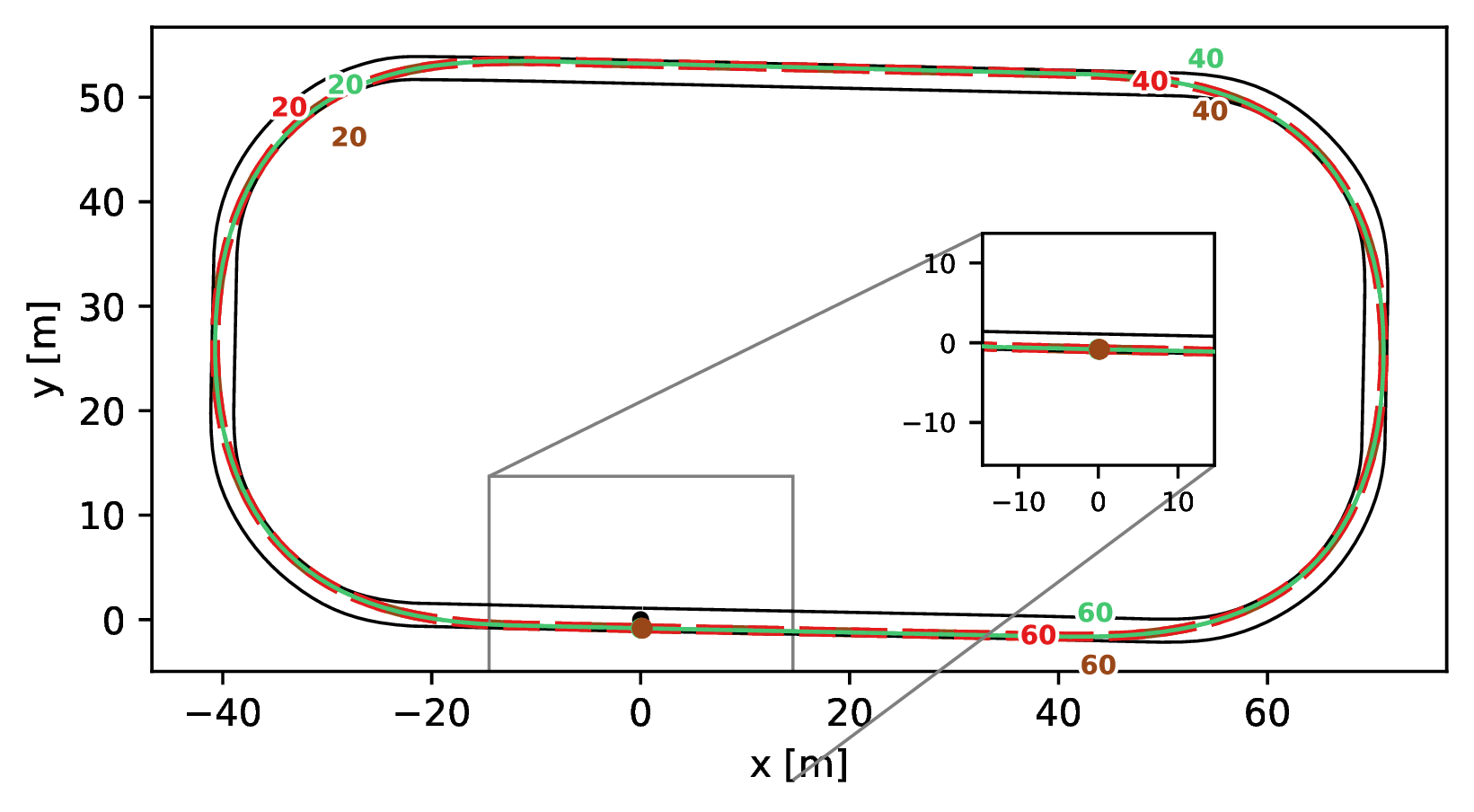}
    \caption{Timestamped closed-loop trajectory comparison between \ddm{}, \fthd{} and \method{} on Austin track (top) and IMS track (bottom). 
    }
    \label{fig:sota_timestamp_austin}
\end{figure}
Fig.~\ref{fig:closed_loop_sota_diagnostics} provides additional insight into the ETHZMobil rollout. While all 3 \ddm{}, \fthd{} and \method{} closely follow the reference raceline, \method{} maintains higher speed through several curved sections, leading to a shorter overall lap time. 
The associated state trajectories remain bounded throughout the rollout, and the predicted tire-force profiles remain physically reasonable, indicating stable recursive use of the learned dynamics model within NMPC.

Additional closed-loop trajectory comparisons for the two additional unseen tracks namely Austin and IMS are shown in Fig.~\ref{fig:sota_timestamp_austin}. 
Although all models remain feasible on these circuits, the timestamped trajectories reveal that \method{} reaches equivalent track locations earlier in the rollout, consistent with the reduced lap times reported in Table~\ref{tab:closed_loop_sota}. 
The zoomed insets further show that all methods follow broadly similar racing lines, with local deviations emerging in curved sections where small differences in the predicted dynamics influence the controller-selected trajectory.

Overall, the closed-loop results demonstrate that \method{} provides a more effective internal prediction model for NMPC than the evaluated \sota{} baselines. 
Notably, these improvements are achieved without uniformly attaining the lowest error across every open-loop metric, highlighting that closed-loop controller performance depends not only on prediction accuracy, but also on the consistency and reliability of the learned dynamics model when repeatedly evaluated within the optimization horizon.
\section{Conclusion}\label{sec:conclusion}
This paper presented \method{}, a physics-aware state-space memory framework for vehicle dynamics learning in autonomous racing. 
The proposed architecture combines state-space and recurrent temporal representations with bounded parameter adaptation and residual correction to capture both long-term operating context and short-term transient behavior while preserving a structured physics-based model. 
Experimental results on simulated and full-scale autonomous racecar datasets demonstrated that \method{} improves prediction accuracy while maintaining low computational complexity and real-time inference capability. 
Open-loop and closed-loop ablation studies further showed that temporal memory, parameter adaptation, and residual correction each contribute to predictive accuracy and robustness. 
Closed-loop NMPC experiments demonstrated that \method{} remains effective when embedded within the controller optimization loop, improving feasibility and reducing lap times relative to representative state-of-the-art dynamics models.

Overall, the results demonstrate that integrating temporal memory, bounded physics-guided parameter adaptation, and residual correction yields an accurate, interpretable, and control-oriented dynamics model for autonomous racing.

Despite these encouraging results, several limitations remain. 
First, the proposed framework does not explicitly model epistemic or aleatoric uncertainty, resulting in deterministic predictions that may be less informative under highly uncertain operating conditions. 
Second, although \method{} is designed to be computationally efficient, its deployment has not yet been evaluated on embedded computing platforms or resource-constrained automotive hardware. 
Future work will investigate uncertainty-aware extensions of the framework and real-time deployment on embedded autonomous racing systems, with the broader goal of enabling robust learning-based vehicle dynamics models for safety-critical control applications.

\section*{Acknowledgment}
ChatGPT 5.1 [OpenAI] was used  in the preparation to improve the clarity, grammar, and readability of the text. 



\printbibliography

\end{document}